\documentclass{article}

\usepackage[ruled,vlined,linesnumbered]{algorithm2e}
\usepackage[utf8]{inputenc} 
\usepackage[T1]{fontenc}    
\usepackage{hyperref}       
\usepackage{url}            
\usepackage{booktabs}       
\usepackage{amsfonts}       
\usepackage{nicefrac}
\usepackage{amsthm}
\usepackage{graphicx}
\usepackage{amsmath,amssymb}
\usepackage{enumitem}
\usepackage{xcolor}
\usepackage{array}
\usepackage{longtable}
\usepackage{subfigure}
\newcommand{\T}{\mathcal{T}}
\DeclareMathOperator*{\argmax}{arg\,max}
\DeclareMathOperator*{\argmin}{arg\,min}
\DeclareMathOperator{\sign}{sign}

\usepackage[accepted]{icml2020}

\icmltitlerunning{Agent57: Outperforming the Atari Human Benchmark}

\begin{document}

\twocolumn[
\icmltitle{Agent57: Outperforming the Atari Human Benchmark}



\icmlsetsymbol{equal}{*}

\begin{icmlauthorlist}
\icmlauthor{Adri\`a Puigdom\`enech Badia}{equal,to}
\icmlauthor{Bilal Piot}{equal,to}
\icmlauthor{Steven Kapturowski}{equal,to}
\icmlauthor{Pablo Sprechmann}{equal,to}
\icmlauthor{Alex Vitvitskyi}{to}
\icmlauthor{Daniel Guo}{to}
\icmlauthor{Charles Blundell}{to}
\end{icmlauthorlist}

\icmlaffiliation{to}{DeepMind}
\icmlcorrespondingauthor{Adri\`a Puigdom\`enech Badia}{adriap@google.com}

\icmlkeywords{Machine Learning, Deep Reinforcement Learning, Exploration}

\vskip 0.3in
]



\printAffiliationsAndNotice{\icmlEqualContribution} 

\begin{abstract}
Atari games have been a long-standing benchmark in the reinforcement learning (RL) community for the past decade. This benchmark was proposed to test general competency of RL algorithms. 
Previous work has achieved good average performance by doing outstandingly well on many games of the set, but very poorly in several of the most challenging games. We propose Agent57, the first deep RL agent that outperforms the standard human benchmark on all 57 Atari games. To achieve this result, we train a neural network which parameterizes a family of policies ranging from very exploratory to purely exploitative. We propose an adaptive mechanism to choose which policy to prioritize throughout the training process. Additionally, we utilize a novel parameterization of the architecture that allows for more consistent and stable learning.
\end{abstract}

\section{Introduction}
\label{sec:introduction}
The Arcade Learning Environment~\citep[ALE; ][]{bellemare2012arcade} was proposed as a platform for empirically assessing agents designed for general competency across a wide range of games. 
ALE offers an interface to a diverse set of Atari 2600 game environments designed to be engaging and challenging for human players.
As~\citet{bellemare2012arcade} put it, the Atari 2600 games are well suited for evaluating general competency in AI agents for three main reasons: \emph{(i)} varied enough to claim generality, \emph{(ii)} each interesting enough to be representative of settings that might be faced in practice, and \emph{(iii)} each created by an independent party to be free of experimenter’s bias.

Agents are expected to perform well in as many games as possible making minimal assumptions about the domain at hand and without the use of game-specific information.
Deep Q-Networks \citep[DQN ;][]{mnih2015human} was the first algorithm to achieve human-level control in
a large number of the Atari 2600 games, measured by human normalized scores (HNS). Subsequently, using HNS to assess performance on Atari games has become one of the most widely used benchmarks in deep reinforcement learning (RL), despite the human baseline scores potentially under-estimating human performance relative to what is possible~\citep{toromanoff2019deep}.
Nonetheless, human benchmark performance remains an oracle for ``reasonable performance'' across the 57 Atari games.
Despite all efforts, no single RL algorithm has been able to achieve over 100\% HNS on all 57 Atari games with one set of hyperparameters. 
Indeed, state of the art algorithms in model-based RL, MuZero~\citep{schrittwieser2019mastering}, and in model-free RL, R2D2~\citep{kapturowski2018recurrent} surpass 100\% HNS on 51 and 52 games, respectively. 
\begin{figure}[!t]
    \centering
    \includegraphics[width=0.45\textwidth]{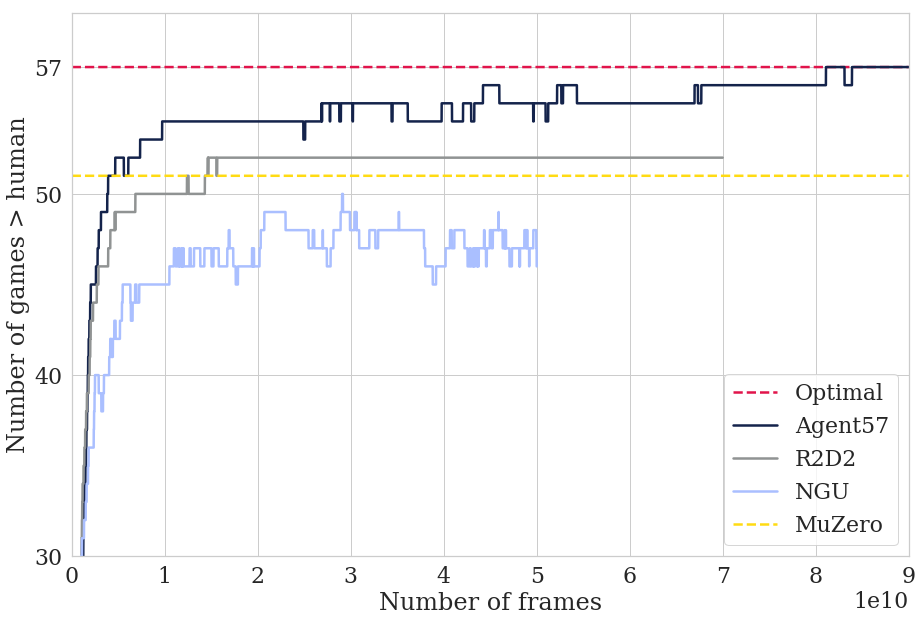}
    \caption{Number of games where algorithms are better than the human benchmark throughout training for Agent57 and state-of-the-art baselines on the 57 Atari games.} 
    \label{fig:cappedhns}
    \vspace{-2ex}
\end{figure}
While these algorithms achieve well above average human-level performance on a large fraction of the games (e.g. achieving more than 1000\% HNS), in the games they fail to do so, they often fail to learn completely.
These games showcase particularly important issues that a general RL algorithm should be able to tackle. Firstly, \emph{long-term credit assignment}: which decisions are most deserving of credit for the positive (or negative) outcomes that follow? This problem is particularly hard when rewards are delayed and credit needs to be assigned over long sequences of actions, such as in the games of \textit{Skiing} or \textit{Solaris}. The game of \textit{Skiing} is a canonical example due to its peculiar reward structure. The goal of the game is to run downhill through all gates as fast as possible. A penalty of five seconds is given for each missed gate.  The reward, given only at the end, is proportional to the time elapsed. Therefore long-term credit assignment is needed to understand why an action taken early in the game (e.g. missing a gate) has a negative impact in the obtained reward. 
Secondly, \emph{exploration}: efficient exploration can be critical to effective learning in RL. 
Games like \textit{Private Eye}, \textit{Montezuma's Revenge}, \textit{Pitfall!} or \textit{Venture} are widely considered hard exploration games~\citep{bellemare2016unifying, ostrovski2017count} as hundreds of actions may be required before a first positive reward is seen. In order to succeed, the agents need to keep exploring the environment despite the apparent impossibility of finding positive rewards.
These problems are particularly challenging in large high dimensional state spaces where function approximation is required.

Exploration algorithms in deep RL generally fall into three categories:
randomized value functions \citep{osband2016deep,fortunato2017noisy,salimans2017evolution,plappert2017parameter,osband2018randomized}, unsupervised policy learning \citep{gregor2016variational,achiam2018variational,eysenbach2018diversity} and intrinsic motivation \citep{schmidhuber1991possibility,oudeyer2007intrinsic,barto2013intrinsic,bellemare2016unifying,ostrovski2017count,fu2017ex2,tang2017exploration,burda2018exploration,choi2018contingency,savinov2018episodic, badia2020never}. Other work combines handcrafted features, domain-specific knowledge or privileged pre-training to side-step the exploration problem, sometimes only evaluating on a few Atari games \citep{aytar2018playing,ecoffet2019go}.
Despite the encouraging results, no algorithm has been able to significantly improve performance on challenging games without deteriorating performance on the remaining games without relying on human demonstrations~\citep{pohlen2018observe}.
Notably, amongst all this work, intrinsic motivation, and in particular, \textit{Never Give Up}~\citep[NGU; ][]{badia2020never} has shown significant recent promise in improving performance on hard exploration games.
NGU achieves this by augmenting the reward signal with an internally generated intrinsic reward that is sensitive to novelty at two levels: short-term novelty within an episode and long-term novelty across episodes.
It then learns a family of policies for exploring and exploiting (sharing the same parameters), with the end goal of obtain the highest score under the exploitative policy.
However, NGU is not the most general agent: much like R2D2 and MuZero are able to perform strongly on all but few games, so too NGU suffers in that it performs strongly on a smaller, \emph{different} set of games to agents such as MuZero and R2D2 (despite being based on R2D2).
For example, in the game  \textit{Surround} R2D2 achieves the optimal score while NGU performs similar to a random policy.
One shortcoming of NGU is that it collects the same amount of experience following each of its policies, regardless of their contribution to the learning progress.
Some games require a significantly different degree of exploration to others.
Intuitively, one would want to allocate the shared resources (both network capacity and data collection) such that end performance is maximized.
We propose allowing NGU to adapt its exploration strategy over the course of an agent's lifetime, enabling specialization to the particular game it is learning.
This is the first significant improvement we make to NGU to allow it to be a more general agent.

Recent work on long-term credit assignment can be categorized into roughly two types: ensuring that gradients correctly assign credit \citep{ke2017sparse,weber2019credit,ferret2019credit,fortunato2019generalization} and
using values or targets to ensure correct credit is assigned \citep{arjona2019rudder,hung2019optimizing,liu2019sequence,harutyunyan2019hindsight}.
NGU is also unable to cope with long-term credit assignment problems such as \textit{Skiing} or \textit{Solaris} where it fails to reach 100\% HNS.
Advances in credit assignment in RL often involve a mixture of both approaches, as values and rewards form the loss whilst the flow of gradients through a model directs learning.

In this work, we propose tackling the long-term credit assignment problem by improving the overall training stability, dynamically adjusting the discount factor, and increasing the backprop through time window.
These are relatively simple changes compared to the approaches proposed in previous work, but we find them to be effective.
Much recent work has explored this problem of how to dynamically adjust hyperparameters of a deep RL agent, e.g., approaches based upon evolution~\citep{jaderberg2017population}, gradients~\citep{xu2018meta} or multi-armed bandits~\citep{schaul2019adapting}.
Inspired by \citet{schaul2019adapting}, we propose using a simple non-stationary multi-armed bandit~\citep{garivier2008upperconfidence} to directly control the exploration rate and discount factor to maximize the episode return, and then provide this information to the value network of the agent as an input.
Unlike~\citet{schaul2019adapting}, 1) it controls the exploration rate and discount factor (helping with long-term credit assignment), and
2) the bandit controls a family of state-action value functions that back up the effects of exploration and longer discounts, rather than linearly tilting a common value function by a fixed functional form.

In summary, our contributions are as follows:
\begin{enumerate}[leftmargin=12pt, align=left, labelwidth=10pt,  labelsep=0pt]
    \item A new parameterization of the state-action value function that decomposes the contributions of the intrinsic and extrinsic rewards. As a result, we significantly increase the training stability over a large range of intrinsic reward scales.
    \item A \emph{meta-controller}: an adaptive mechanism to select which of the policies (parameterized by exploration rate and discount factors) to prioritize throughout the training process. This allows the agent to control the \textit{exploration/exploitation trade-off} by dedicating more resources to one or the other.
    \item Finally, we demonstrate for the first time performance that is above the human baseline across all Atari 57 games. As part of these experiments, we also find that simply re-tuning the backprop through time window to be twice the previously published window for R2D2 led to superior long-term credit assignment (e.g., in \textit{Solaris}) while still maintaining or improving overall performance on the remaining games.
\end{enumerate}

These improvements to NGU collectively transform it into the most general Atari 57 agent, enabling it to outperform the human baseline uniformly over all Atari 57 games. Thus, we call this agent: Agent57.

\section{Background: Never Give Up (NGU)}
\label{sec:background}
Our work builds on top of the NGU agent, which combines two ideas: first, the curiosity-driven exploration, and second, distributed deep RL agents, in particular R2D2.

NGU computes an intrinsic reward in order to encourage exploration. This reward is defined by combining per-episode and life-long novelty. The per-episode novelty, $r_t^{\text{episodic}}$, rapidly vanishes over the course of an episode, and it is computed by comparing observations to the contents of an episodic memory. The life-long novelty, $\alpha_t$, slowly vanishes throughout training, and it is computed by using a parametric model (in NGU and in this work Random Network Distillation~\citep{burda2018exploration} is used to this end). With this, the intrinsic reward $r^i_t$ is defined as follows:
\begin{equation*}
r_t^{i} = r_t^{\text{episodic}}\cdot \min \left\{\max\left\{\alpha_t, 1\right\}, L\right\},
\label{eq:clipping}
\end{equation*}
where $L=5$ is a chosen maximum reward scaling. This leverages the long-term novelty provided by $\alpha_t$, while $r_t^{\text{episodic}}$ continues to encourage the agent to explore within an episode. For a detailed description of the computation of $r_t^{\text{episodic}}$ and $\alpha_t$, see~\citep{badia2020never}.
At time $t$, NGU adds $N$ different scales of the same intrinsic reward $\beta_j r_t^i$ ($\beta_j\in\mathbb{R}^+$, $j\in 0,\dots N-1$) to the extrinsic reward provided by the environment, $r^e_t$, to form $N$ potential total rewards $r_{j,t} = r^e_t + \beta_j r_t^i$. Consequently, NGU aims to learn the $N$ different associated optimal state-action value functions $Q^*_{r_j}$ associated with each reward function $r_{j,t}$. The exploration rates $\beta_j$ are parameters that control the degree of exploration. Higher values will encourage exploratory policies and smaller values will encourage exploitative policies.
Additionally, for purposes of learning long-term credit assignment, each $Q^*_{r_j}$ has its own associated discount factor $\gamma_j$ (for background and notations on Markov Decision Processes (MDP) see App.~\ref{app:background}). Since the intrinsic reward is typically much more dense than the extrinsic reward, $\{(\beta_j, \gamma_j)\}_{j=0}^{N-1}$ are chosen so as to allow for long term horizons (high values of $\gamma_j$) for exploitative policies (small values of $\beta_j$) and small term horizons (low values of $\gamma_j$) for exploratory policies (high values of $\beta_j$).

To learn the state-action value function $Q^*_{r_j}$, NGU trains a recurrent neural network $Q(x, a, j; \theta)$, where $j$ is a one-hot vector indexing one of $N$ implied MDPs (in particular $(\beta_j, \gamma_j)$), $x$ is the current observation, $a$ is an action, and $\theta$ are the parameters of the network (including the recurrent state).
In practice, NGU can be unstable and fail to learn an appropriate approximation of $Q^*_{r_j}$ for all the state-action value functions in the family, even in simple environments. This is especially the case when the scale and sparseness of $r^e_t$ and $r^i_t$ are both different, or when one reward is more noisy than the other. We conjecture that learning a common state-action value function for a mix of rewards is difficult when the rewards are very different in nature. Therefore, in Sec.~\ref{subsec:value1}, we propose an architectural modification to tackle this issue.

\begin{figure}[!ht]
    \centering
    \includegraphics[width=0.4\textwidth]{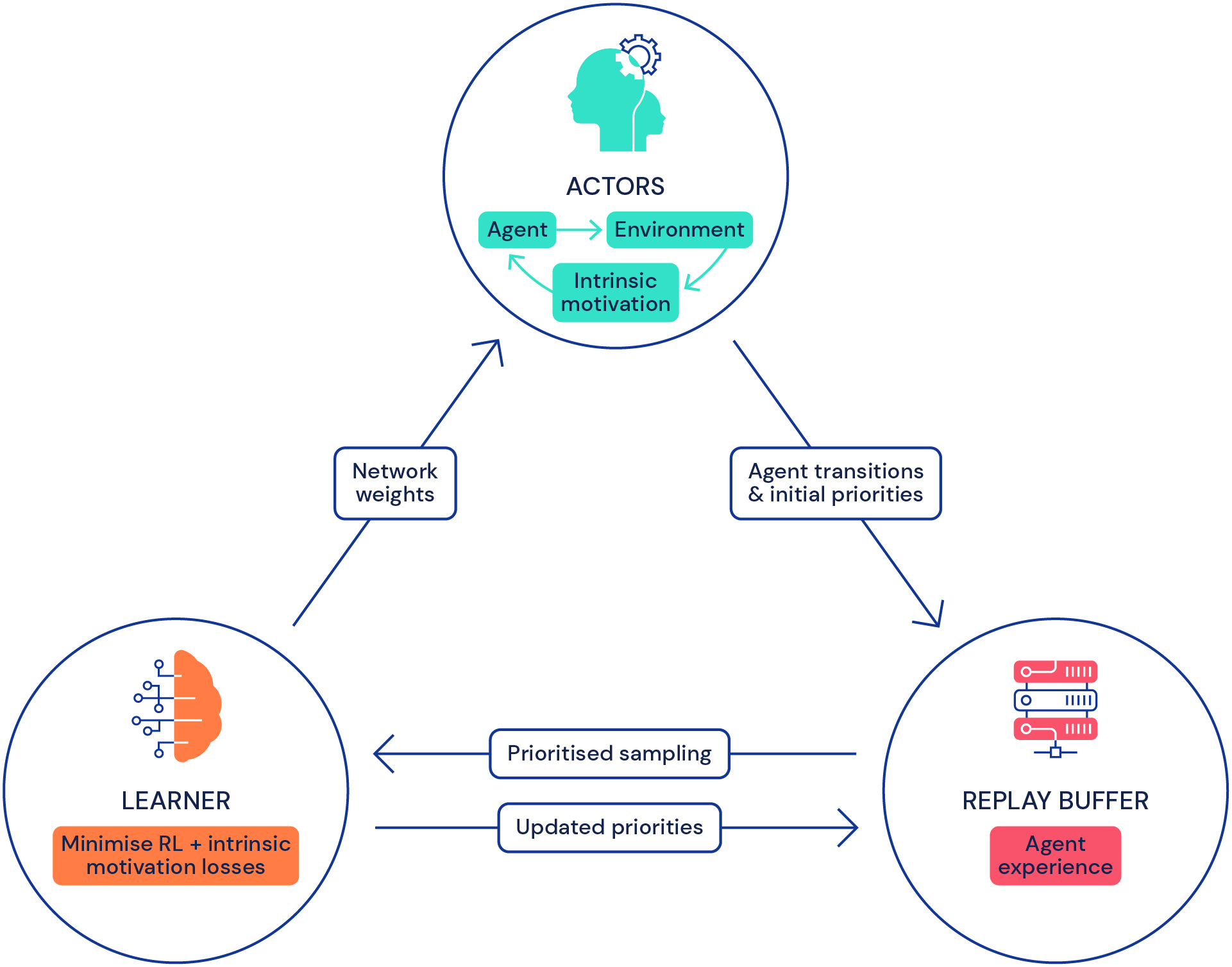}
    \caption{A schematic depiction of a distributed deep RL agent.} 
    \label{fig:distributed}
\end{figure}

Our agent is a deep distributed RL agent, in the lineage of R2D2 and NGU. As such, it decouples the data collection and the learning processes by having many actors feed data to a central prioritized replay buffer. A learner can then sample training data from this buffer, as shown in Fig.~\ref{fig:distributed} (for implementation details and hyperparameters refer to App.~\ref{app:distributed}). 
More precisely, the replay buffer contains sequences of transitions that are removed regularly in a FIFO-manner. These sequences come from actor processes that interact with independent copies of the environment, and they are prioritized based on temporal differences errors~\citep{kapturowski2018recurrent}. The priorities are initialized by the actors and updated by the learner with the updated state-action value function $Q(x, a, j; \theta)$. According to those priorities, the learner samples sequences of transitions from the replay buffer to construct an RL loss. Then, it updates the parameters of the neural network $Q(x, a, j; \theta)$ by minimizing the RL loss to approximate the optimal state-action value function. Finally, each actor shares the same network architecture as the learner but with different weights. We refer as $\theta_l$ to the parameters of the $l-$th actor. The learner weights $\theta$ are sent to the actor frequently, which allows it to update its own weights $\theta_l$. Each actor uses different values $\epsilon_l$, which are employed to follow an $\epsilon_l$-greedy policy based on the current estimate of the state-action value function $Q(x, a, j; \theta_l)$. In particular, at the beginning of each episode and in each actor, NGU uniformly selects a pair $(\beta_j, \gamma_j)$. We hypothesize that this process is sub-optimal and propose to improve it in Sec.~\ref{subsec:adaptive1} by introducing a meta-controller for each actor that adapts the data collection process. 

\section{Improvements to NGU}
\label{sec:improvments}

\subsection{State-Action Value Function Parameterization}
\label{subsec:value1}

The proposed architectural improvement consists in splitting the state-action value function in the following way:
\begin{equation*}
Q(x, a, j; \theta) = Q(x, a, j; \theta^e) + \beta_jQ(x, a, j; \theta^i),
\end{equation*}
where $Q(x, a, j; \theta^e)$ and $Q(x, a, j; \theta^i)$ are the extrinsic and intrinsic components of $Q(x, a, j; \theta)$ respectively. The sets of weights $\theta^e$ and $\theta^i$ separately parameterize two neural networks with identical architecture and $\theta = \theta^i\cup\theta^e$. Both $Q(x, a, j; \theta^e)$ and $Q(x, a, j; \theta^i)$ are optimized separately in the learner with rewards $r^e$ and $r^i$ respectively, but with the same target policy $\pi(x) = \argmax_{a\in\mathcal{A}}Q(x, a, j; \theta)$. More precisely, to train the weights $\theta^e$ and $\theta^i$, we use the same sequence of transitions sampled from the replay, but with two different transformed Retrace loss functions~\citep{munos2016safe}. For $Q(x, a, j; \theta^e)$ we compute an extrinsic transformed Retrace loss on the sequence transitions with rewards $r^e$ and target policy $\pi$, whereas for $Q(x, a, j; \theta^i)$ we compute an intrinsic transformed Retrace loss on the same sequence of transitions but with rewards $r^i$ and target policy $\pi$. A reminder of how to compute a transformed Retrace loss on a sequence of transitions with rewards $r$ and target policy $\pi$ is provided in App.~\ref{app:retrace}.

In addition, in App.~\ref{app:decoupling}, we show that this optimization of separate state-action values is equivalent to the optimization of the original single state-action value function with reward $r^e + \beta_jr^i$ (under a simple gradient descent optimizer). Even though the theoretical objective being optimized is the same, the parameterization is different: we use two different neural networks to approximate each one of these state-action values (a schematic and detailed figures of the architectures used can be found in App.~\ref{app:neural}). By doing this, we allow each network to adapt to the scale and variance associated with their corresponding reward, and we also allow for the associated optimizer state to be separated for intrinsic and extrinsic state-action value functions.

Moreover, when a transformed Bellman operator~\citep{pohlen2018observe} with function $h$ is used (see App.~\ref{app:background}), we can split the state-action value function in the following way:
\begin{align*}
&Q(x, a, j; \theta)= 
\\
&h\left(h^{-1}(Q(x, a,j; \theta^e)) + \beta_j h^{-1}(Q(x, a,j; \theta^i))\right).
\end{align*}
In App.~\ref{app:decoupling}, we also show that the optimization of separated transformed state-action value functions is equivalent to the optimization of the original single transformed state-action value function. In practice, choosing a simple or transformed split does not seem to play an important role in terms of performance (empirical evidence and an intuition behind this result can be found in App.~\ref{app:mix}). In our experiments, we choose an architecture with a simple split which corresponds to $h$ being the identity, but still use the transformed Retrace loss functions.

\subsection{Adaptive Exploration over a Family of Policies}
\label{subsec:adaptive1}
The core idea of NGU is to jointly train a family of policies with different degrees of exploratory behaviour using a single network architecture.
In this way, training these exploratory policies plays the role of a set of auxiliary tasks that can help train the shared architecture even in the absence of extrinsic rewards.
A major limitation of this approach is that all policies are trained equally, regardless of their contribution to the learning progress. 
We propose to incorporate a meta-controller that can adaptively select which policies to use both at training and evaluation time.
This carries two important consequences. Firstly, by selecting which policies to prioritize during training, we can allocate more of the capacity of the network to better represent the state-action value function of the policies that are most relevant for the task at hand.
Note that this is likely to change throughout the training process, naturally building a curriculum to facilitate training.
As mentioned in Sec.~\ref{sec:background}, policies are represented by pairs of exploration rate and discount factor, $(\beta_j, \gamma_j)$, which determine the discounted cumulative rewards to maximize. It is natural to expect policies with higher $\beta_j$ and lower $\gamma_j$ to make more progress early in training, while the opposite would be expected as training progresses.
Secondly, this mechanism also provides a natural way of choosing the best policy in the family to use at evaluation time.
Considering a wide range of values of $\gamma_j$ with $\beta_j \approx 0$, provides a way of automatically adjusting the discount factor on a per-task basis. This significantly increases the generality of the approach.

We propose to implement the meta-controller using a non-stationary multi-arm bandit algorithm running independently on each actor. The reason for this choice, as opposed to a global meta-controller, is that each actor follows a different $\epsilon_l$-greedy policy which may alter the choice of the optimal arm.
Each arm $j$ from the $N$-arm bandit is linked to a policy in the family and corresponds to a pair $(\beta_j, \gamma_j)$.
At the beginning of each episode, say, the $k$-th episode, the meta-controller chooses an arm $J_k$ setting which policy will be executed. We use capital letters for the arm $J_k$ because it is a random variable.
Then the $l$-th actor acts $\epsilon_l$-greedily with respect to the corresponding state-action value function, $Q(x, a, J_k; \theta_l)$, for the whole episode. 
The undiscounted extrinsic episode returns, noted $R^e_k(J_k)$, are used as a reward signal to train the multi-arm bandit algorithm of the meta-controller. 

The reward signal $R^e_k(J_k)$ is non-stationary, as the agent changes throughout training. 
Thus, a classical bandit algorithm such as Upper Confidence Bound \citep[UCB; ][]{garivier2008upperconfidence} will not be able to adapt to the changes of the reward through time. Therefore, we employ a simplified sliding-window UCB with $\epsilon_{\texttt{UCB}}$-greedy exploration.
With probability $1-\epsilon_{\texttt{UCB}}$, this algorithm runs a slight modification of classic UCB on a sliding window of size $\tau$ and selects a random arm with probability $\epsilon_{\texttt{UCB}}$ (details of the algorithms are provided in App.~\ref{app:bandits}). 

\begin{figure}[!t]
    \centering
    \includegraphics[width=0.5\textwidth]{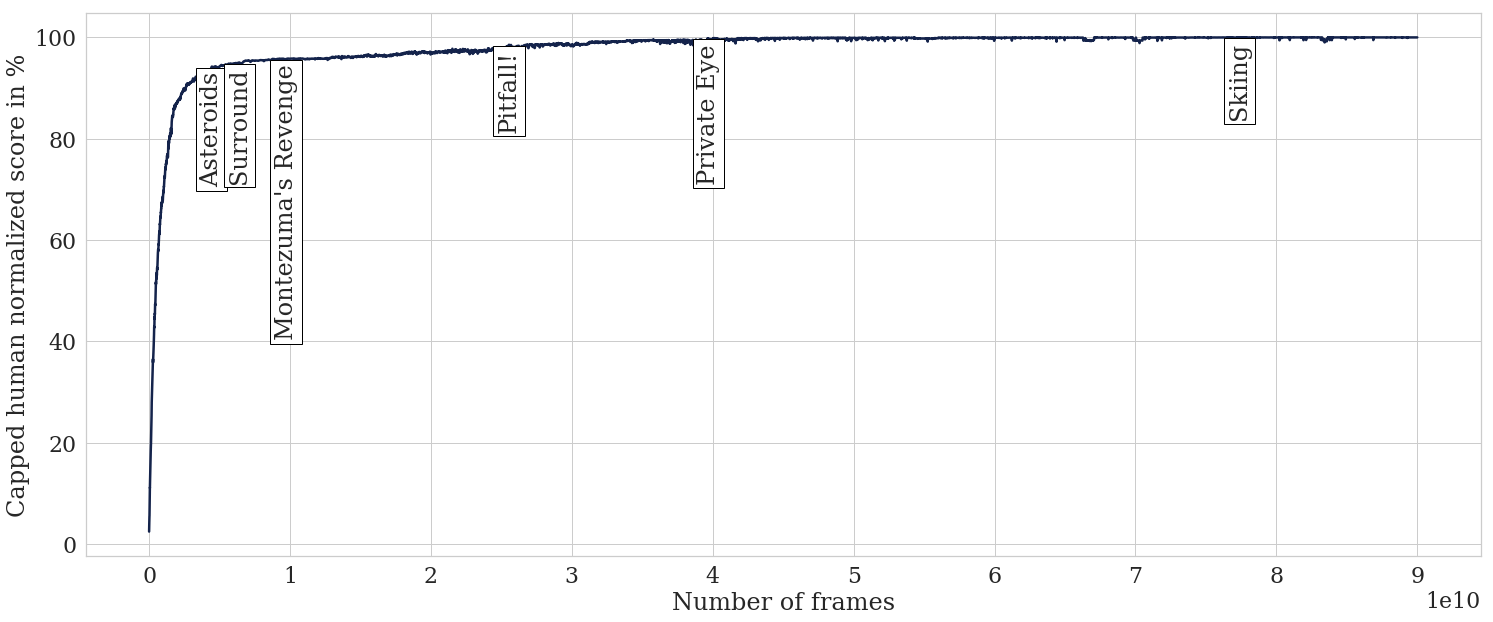}
    \vspace{-3ex}
    \caption{Capped human normalized score where we observe at which point the agent surpasses the human benchmark on the last 6 games.} 
    \label{fig:superhuman_timeline}
    \vspace{-3ex}
\end{figure}

Note that the benefit of adjusting the discount factor through training and at evaluation could be applied even in the absence of intrinsic rewards. To show this, we propose augmenting a variant of R2D2 with a meta-controller. In order to isolate the contribution of this change, we evaluate a variant of R2D2 which uses the same RL loss as Agent57.
Namely, a transformed Retrace loss as opposed to a transformed n-step loss as in the original paper. We refer to this variant as R2D2 (Retrace) throughout the paper. In all other aspects, R2D2 (Retrace) is exactly the same algorithm as R2D2.
We incorporate the joint training of several policies parameterized by $\{\gamma_j\}_{j=0}^{N-1}$ to R2D2 (Retrace). We refer to this algorithm as \emph{R2D2 (bandit)}.

\section{Experiments}
\label{sec:experiments}

We begin this section by describing our experimental setup. Following NGU, Agent57 uses a family of coefficients $\{(\beta_j,\gamma_j)\}_{j=0}^{N-1}$ of size $N=32$. The choice of discounts $\{\gamma_j\}_{j=0}^{N-1}$ differs from that of NGU to allow for higher values, ranging from $0.99$ to $0.9999$ (see App.~\ref{app:family} for details). The meta-controller uses a window size of $\tau=160$ episodes and $\epsilon=0.5$ for the actors and a window size of $\tau=3600$ episodes and $\epsilon=0.01$. All the other hyperparameters are identical to those of NGU, including the standard preprocessing of Atari frames. For a complete description of the hyperparameters and preprocessing we use, please see App.~\ref{app:hyperparameters}. For all agents we run (that is, all agents except MuZero where we report numbers presented in~\citet{schrittwieser2019mastering}), we employ a separate evaluator process to continuously record scores. We record the undiscounted episode returns averaged over $3$ seeds and using a windowed mean over $50$ episodes. For our best algorithm, Agent57, we report the results averaged over $6$ seeds on all games to strengthen the significance of the results. On that average, we report the maximum over training as their final score, as done in~\citet{fortunato2017noisy, badia2020never}. Further details on our evaluation setup are described in App.~\ref{app:distributed}.

In addition to using human normalized scores $\text{HNS}=\frac{\text{Agent}_{\text{score}}-\text{Random}_{\text{score}}}{\text{Human}_{\text{score}}-\text{Random}_{\text{score}}}$, we report the capped human normalized scores, $\text{CHNS} = \max\{\min\{\text{HNS}, 1\}, 0\}$.
This measure is a better descriptor for evaluating general performance, as it puts an emphasis in the games that are below the average human performance benchmark. Furthermore, and avoiding any issues that aggregated metrics may have, we also provide all the scores that all the ablations obtain in all games we evaluate in App.~\ref{app:tabatari10}.

\begin{table*}
\scriptsize
\centering
\caption{Number of games above human, mean capped, mean and median human normalized scores for the 57 Atari games.}
\vspace{1ex}
\begin{tabular}{|c|c|c|c|c|c|c|}
\hline
 Statistics & Agent57 & R2D2 (bandit) & NGU & R2D2 (Retrace) & R2D2 & MuZero \\
\hline
 Capped mean & \bf{100.00} & 96.93 & 95.07 & 94.20 & 94.33 & 89.92 \\
 Number of games $>$ human & \bf{57} & 54 & 51 & 52 & 52 & 51 \\
 Mean & 4766.25 & 5461.66 & 3421.80 & 3518.36 & 4622.09 & \bf{5661.84} \\
 Median & 1933.49 & 2357.92 & 1359.78 & 1457.63 & 1935.86 & \bf{2381.51} \\
 40th Percentile & 1091.07 & \bf{1298.80} & 610.44 & 817.77 & 1176.05 & 1172.90 \\
 30th Percentile & 614.65 & \bf{648.17} & 267.10 & 420.67 & 529.23 & 503.05 \\
 20th Percentile & \bf{324.78} & 303.61 & 226.43 & 267.25 & 215.31 & 171.39 \\
 10th Percentile & \bf{184.35} & 116.82 & 107.78 & 116.03 & 115.33 & 75.74 \\
 5th Percentile & \bf{116.67} & 93.25 & 64.10 & 48.32 & 50.27 & 0.03 \\
\hline
\end{tabular}
\label{tab:normalizedscores}
\end{table*}

We structure the rest of this section in the following way: firstly, we show an overview of the results that Agent57 achieves. Then we proceed to perform ablations on each one of the improvements we propose for our model.

\subsection{Summary of the Results}
\label{subsec:summary}

Tab.~\ref{tab:normalizedscores} shows a summary of the results we obtain on all 57 Atari games when compared to baselines. MuZero obtains the highest uncapped mean and median human normalized scores, but also the lowest capped scores. This is due to the fact that MuZero performs remarkably well in some games, such as \textit{Beam Rider}, where it shows an uncapped score of $27469\%$, but at the same time catastrophically fails to learn in games such as \textit{Venture}, achieving a score that is on par with a random policy. We see that the meta-controller improvement successfully transfers to R2D2: the proposed variant R2D2 (bandit) shows a mean, median, and CHNS that are much higher than R2D2 with the same Retrace loss. Finally, Agent57 achieves a median and mean that is greater than NGU and R2D2, but also its CHNS is 100\%. This shows the generality of Agent57: not only it obtains a strong mean and median, but also it is able to obtain strong performance on the tail of games in which MuZero and R2D2 catastrophically fail. This is more clearly observed when looking at different percentiles: up to the $20$th percentile, Agent57 shows much greater performance, only slightly surpassed by R2D2 (bandit) when we examine higher percentiles.
In Fig.~\ref{fig:superhuman_timeline} we report the performance of Agent57 in isolation on the $57$ games. We show the last $6$ games (in terms of number of frames collected by the agents) in which the algorithm surpasses the human performance benchmark. As shown, the benchmark over games is beaten in a long-tailed fashion, where Agent57 uses the first $5$ billion frames to surpass the human benchmark on $51$ games. After that, we find hard exploration games, such as \textit{Montezuma's Revenge}, \textit{Pitfall!}, and \textit{Private Eye}. Lastly, Agent57 surpasses the human benchmark on \textit{Skiing} after $78$ billion frames.
To be able to achieve such performance on \textit{Skiing}, Agent57 uses a high discount (as we show in Sec. \ref{subsec:adaptive2}). This naturally leads to high variance in the returns, which leads to needing more data in order to learn to play the game. One thing to note is that, in the game of \textit{Skiing}, the human baseline is very competitive, with a score of  $-4336.9$, where $-17098.1$ is random and $-3272$ is the optimal score one can achieve.

In general, as performance in Atari keeps improving, it seems natural to concentrate on the tail of the distribution, i.e., pay attention to those games for which progress in the literature has been historically much slower than average.
We now present results for a subset of 10 games that we call the \emph{challenging set}. It consists of the six hard exploration games as defined in~\cite{bellemare2016unifying}, plus games that require long-term credit assignment. More concretely, the games we use are: \textit{Beam Rider}, \textit{Freeway}, \textit{Montezuma's Revenge}, \textit{Pitfall!}, \textit{Pong}, \textit{Private Eye}, \textit{Skiing}, \textit{Solaris}, \textit{Surround}, and \textit{Venture}. 

\begin{figure}[!t]
    \centering
    \includegraphics[width=0.47\textwidth]{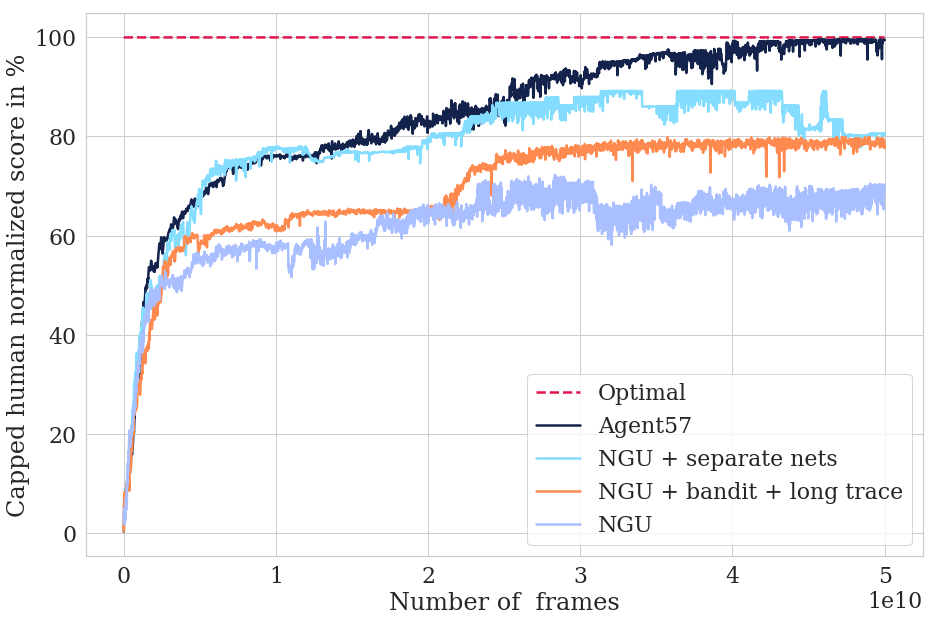}
    \caption{Performance progression on the 10-game \emph{challenging set} obtained from incorporating each one of the improvements.} 
    \label{fig:progression}
\end{figure}

In Fig.~\ref{fig:progression} we can see the performance progression obtained from incorporating each one of the improvements we make on top of NGU. Such performance is reported on the selection of $10$ games mentioned above. We observe that each one of the improvements results in an increment in final performance. Further, we see that each one of the improvements that is part of Agent57 is necessary in order to obtain the consistent final performance of $100\%$ CHNS.

\subsection{State-Action Value Function Parameterization}
\label{subsec:value2}
 
\begin{figure}[!t]
    \centering
    \includegraphics[width=0.45\textwidth]{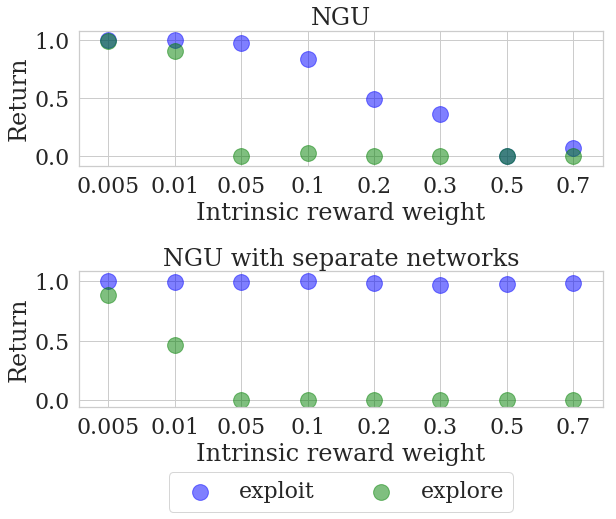}
    \caption{Extrinsic returns for the exploitative ($\beta_0=0$) and most exploratory ($\beta_{31}=\beta$) on ``random coin'' for different values of the intrinsic reward weight, $\beta$. \emph{(Top)} NGU\emph{(Bottom)} NGU with Separate networks for intrinsic and extrinsic values.} 
    \label{fig:randomcoinbar}
    \vspace{-3ex}
\end{figure}

We begin by evaluating the influence of the state-action value function parametrization on a minimalistic gridworld environment, called ``random coin''.
It consists of an empty room of size $15\,\times\,15$ where a coin and an agent are randomly placed at the start of each episode. The agent can take four possible actions (up, down, left right) and episodes are at most $200$ steps long. If the agent steps over the coin, it receives a reward of $1$ and the episode terminates.
In Fig.~\ref{fig:randomcoinbar} we see the results of NGU with and without the new parameterization of its state-action value functions.
We report performance after $150$ million frames. 
We compare the extrinsic returns for the policies that are the exploitative ($\beta_j=0$) and the most exploratory (with the largest $\beta_j$ in the family).
Even for small values of the exploration rates ($\max_j\beta_j$), this setting induces very different exploratory and exploitative policies.
Maximizing the discounted extrinsic returns is achieved by taking the shortest path towards the coin (obtaining an extrinsic return of one), whereas maximizing the augmented returns is achieved by avoiding the coin and visiting all remaining states (obtaining an extrinsic return of zero).
In principle, NGU should be able to learn these policies jointly.
However, we observe that the exploitative policy in NGU struggles to solve the task as intrinsic motivation reward scale increases. 
As we increase the scale of the intrinsic reward, its value becomes much greater than that of the extrinsic reward.
As a consequence, the conditional state-action value network of NGU is required to represent very different values depending on the $\beta_j$ we condition on. This implies that the network is increasingly required to have more flexible representations. 
Using separate networks dramatically increases its robustness to the intrinsic reward weight that is used.
Note that this effect would not occur if the episode did not terminate after collecting the coin. In such case, exploratory and exploitative policies would be allowed to be very similar: both could start by collecting the coin as quickly as possible. 
In Fig.~\ref{fig:progression} we can see that this improvement also translates to the \emph{challenging set}. NGU achieves a much lower average CHNS than its separate network counterpart. We also observe this phenomenon when we incorporate the meta-controller. Agent57 suffers a drop of performance that is greater than $20\%$ when the separate network improvement is removed.

We can also see that it is a general improvement: it does not show worse performance on any of the $10$ games of the challenging set. More concretely, the largest improvement is seen in the case of \textit{Surround}, where NGU obtains a score on par with a random policy, whereas with the new parametrization it reaches a score that is nearly optimal. This is because \textit{Surround} is a case that is similar to the ``random coin'' environment mentioned above: as the player makes progress in the game, they have the choice to surround the opponent snake, receive a reward, and start from the initial state, or keep wandering around without capturing the opponent, and thus visiting new states in the world.

\subsection{Backprop Through Time Window Size}
\label{subsec:tracelength}

In this section we analyze the impact of having a backprop through time window size. More concretely, we analyze its impact on the base algorithm R2D2 to see its effect without NGU or any of the improvements we propose. Further, we also analyze its effect on Agent57, to see if any of the improvements on NGU overlap with this change. In both cases, we compare using backprop through time window sizes of $80$ (default in R2D2) versus $160$.

\begin{figure}[!t]
    \centering
    \includegraphics[width=0.45\textwidth]{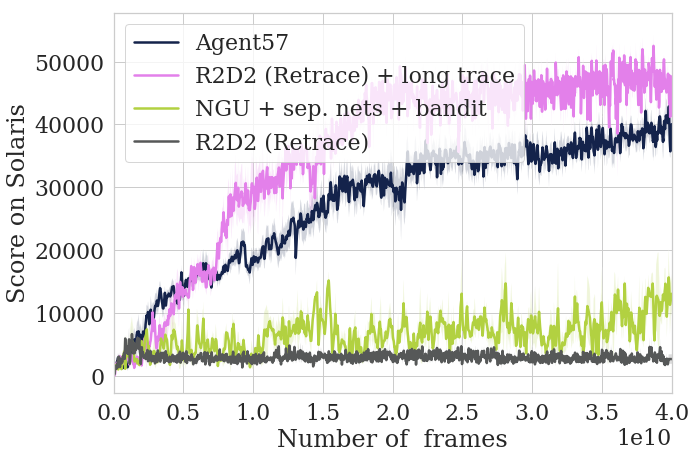}
    \caption{\textit{Solaris} learning curves with small and long backprop through time window sizes for both R2D2 and Agent57.} 
    \label{fig:solaristracelength}
    \vspace{-3ex}
\end{figure}

In aggregated terms over the \emph{challenging set}, its effect seems to be the same for both R2D2 and Agent57: using a longer backprop through time window appears to be initially slower, but results in better overall stability and slightly higher final score. A detailed comparison over those $10$ games is shown in App.~\ref{app:tracelength}. This effect can be seen clearly in the game of \textit{Solaris}, as observed in Fig.~\ref{fig:solaristracelength}. This is also the game showing the largest improvement in terms of final score.
This is again general improvement, as it enhances performance on all the \emph{challenging set} games. For further details we report the scores in App. \ref{app:tabatari10}.

\subsection{Adaptive Exploration}
\label{subsec:adaptive2}
\begin{figure}[!t]
    \centering
    \includegraphics[width=0.45\textwidth]{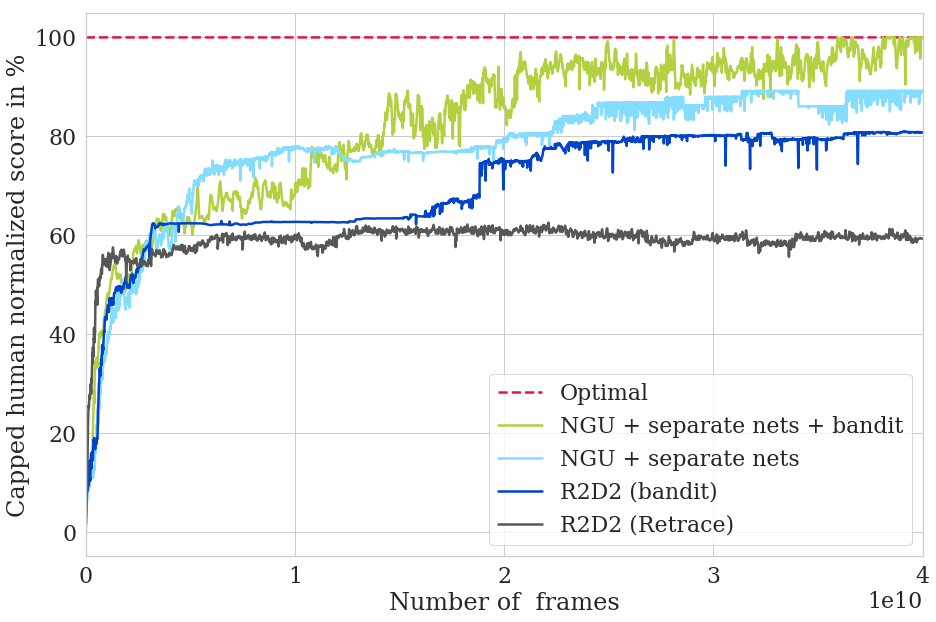}
    \caption{Performance comparison for adaptive exploration on the 10-game \emph{challenging set}.} 
    \label{fig:r2d2progression}
    \vspace{-3ex}
\end{figure}

In this section, we analyze the effect of using the meta-controller described in Sec.~\ref{subsec:value1} in both the actors and the evaluator. To isolate the contribution of this improvement, we evaluate two settings: R2D2 and NGU with separate networks, with and without meta-controller.
Results are shown in Fig. \ref{fig:r2d2progression}. Again, we observe that this is a general improvement in both comparisons. Firstly, we observe that there is a great value in this improvement on its own, enhancing the final performance of R2D2 by close to $20\%$ CHNS. Secondly, we observe that the benefit on NGU with separate networks is more modest than for R2D2. This indicates that there is a slight overlap in the contributions of the separate network parameterization and the use of the meta-controller.
The bandit algorithm can adaptively decrease the value of $\beta$ when the difference in scale between intrinsic and extrinsic rewards is large. 
Using the meta-controller allows to include very high discount values in the set $\{\gamma_j\}_{j=0}^N$.
Specifically, running R2D2 with a high discount factor, $\gamma=0.9999$ surpasses the human baseline in the game of \textit{Skiing}. However, using that hyperparameter across the full set of games, renders the algorithm very unstable and damages its end performance. All the scores in the \emph{challenging set} for a fixed high discount ($\gamma=0.9999$) variant of R2D2 are reported in App.~\ref{app:tabatari10}.
When using a meta-controller, the algorithm does not need to make this compromise: it can adapt it in a per-task manner.

\begin{figure}[!t]
    \centering
    \includegraphics[width=0.45\textwidth]{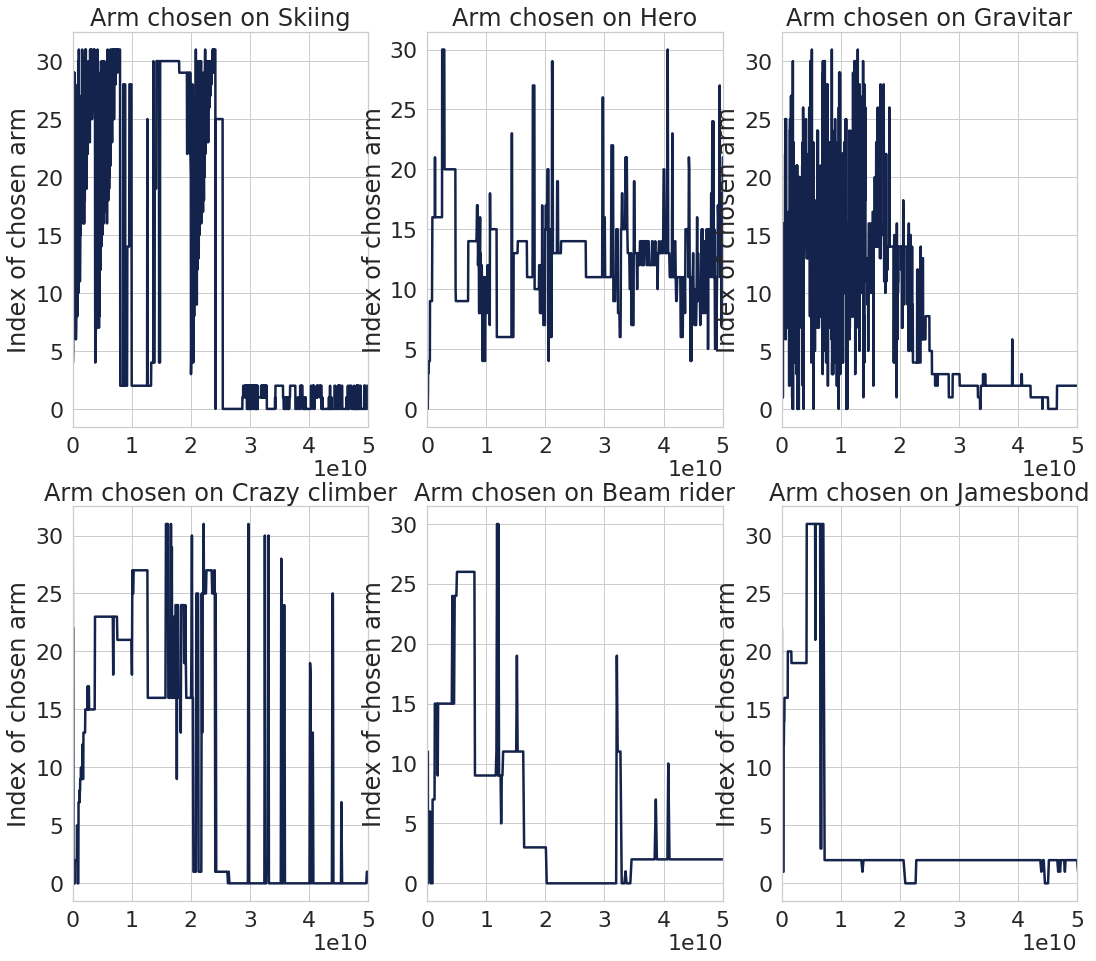}
    \caption{Best arm chosen by the evaluator of Agent57 over training for different games.}
    \label{fig:mixture_chosen}
    \vspace{-3ex}
\end{figure}

Finally, the results and discussion above show why it is beneficial to use different values of $\beta$ and $\gamma$ on a per-task basis. At the same time, in Sec. \ref{sec:improvments} we hypothesize it would also be useful to vary those coefficients throughout training. In Fig. \ref{fig:mixture_chosen} we can see the choice of ($\beta_j$, $\gamma_j$) producing highest returns on the meta-controller of the evaluator across training for several games. Some games clearly have a preferred mode: on \textit{Skiing} the high discount combination is quickly picked up when the agent starts to learn, and on \textit{Hero} a high $\beta$ and low $\gamma$ is generally preferred at all times. On the other hand, some games have different preferred modes throughout training: on \textit{Gravitar}, \textit{Crazy Climber}, \textit{Beam Rider}, and \textit{Jamesbond}, Agent57 initially chooses to focus on exploratory policies with low discount, and, as training progresses, the agent shifts into producing experience from higher discount and more exploitative policies.

\section{Conclusions}
\label{subsec:conclusions}

We present the first deep reinforcement learning agent with performance above the human benchmark on all $57$ Atari games. The agent is able to balance the learning of different skills that are required to be performant on such diverse set of games: \textit{exploration and exploitation} and \textit{long-term credit assignment}. To do that, we propose simple improvements to an existing agent, \textit{Never Give Up}, which has good performance on hard-exploration games, but in itself does not have strong overall performance across all $57$ games. These improvements are i) using a different parameterization of the state-action value function, ii) using a meta-controller to dynamically adapt the novelty preference and discount, and iii) the use of longer backprop-through time window to learn from using the Retrace algorithm.

This method leverages a great amount of computation to its advantage: similarly to NGU, it is able to scale well with increasing amounts of computation. This has also been the case with the many recent achievements in deep RL~\citep{silver2016mastering, andrychowicz2018learning, vinyals2019grandmaster}. While this enables our method to achieve strong performance, an interesting research direction is to pursue ways in which to improve the data efficiency of this agent. Additionally, this agent shows an average capped human normalized score of $100\%$. However, in our view this by no means marks the end of Atari research, not only in terms of efficiency as above, but also in terms of general performance. We offer two views on this: firstly, analyzing the performance among percentiles gives us new insights on how general algorithms are. While Agent57 achieves great results on the first percentiles of the $57$ games and holds better mean and median performance than NGU or R2D2, as MuZero shows, it could still obtain much better average performance. Secondly, as pointed out by~\citet{toromanoff2019deep}, all current algorithms are far from achieving optimal performance in some games. To that end, key improvements to use might be enhancements in the representations that Agent57 and NGU use for exploration, planning (as suggested by the results achieved by MuZero) as well as better mechanisms for credit assignment (as highlighted by the results seen in \textit{Skiing}).

\section*{Acknowledgments}
We thank Daan Wierstra, Koray Kavukcuoglu, Vlad Mnih, Vali Irimia, Georg Ostrovski, Mohammad Gheshlaghi Azar, R\'emi Munos, Bernardo Avila Pires, Florent Altch\'e, Steph Hughes-Fitt, Rory Fitzpatrick, Andrea Banino, Meire Fortunato, Melissa Tan, Benigno Uria, Borja Ibarz, Andre Barreto, Diana Borsa, Simon Osindero, Tom Schaul, and many other colleagues at DeepMind for helpful discussions and comments on the manuscript.

\bibliography{biblio}
\bibliographystyle{icml2020}

\appendix
\onecolumn

\section{Background on MDP}
\label{app:background}
A Markov decision process~\citep[MDP; ][]{puterman1990markov} is a tuple $(\mathcal{X},\mathcal{A}, P, r, \gamma)$, with $\mathcal{X}$ being the state space, $\mathcal{A}$ being the action space, $P$ the state-transition distribution maps each state-action tuple $(x, a)$ to a probability distribution over states (with $P(y|x,a)$ denoting the probability of transitioning to state $y$ from $x$ by choosing action $a$), the reward function $r\in\mathbb{R}^{\mathcal{X}\times\mathcal{A}}$ and $\gamma\in]0, 1[$ the discount factor. A stochastic policy $\pi$ maps each state to a distribution over actions ($\pi(a|x)$ denotes the probability of choosing action $a$ in state $x$). A deterministic policy $\pi_D\in\mathcal{X}^\mathcal{A}$ can also be represented by a distribution over actions $\pi$ such that $\pi(\pi_D(x)|x)=1$. We will use one or the other concept with the same notation $\pi$ in the remaining when the context is clear. 

Let $\T(x, a, \pi)$ be the distribution over trajectories $\tau=(X_t, A_t, R_t, X_{t+1})_{t\in\mathbb{N}}$ generated by a policy $\pi$, with $(X_0,A_0)=(x,a)$, $\forall t\geq1, A_t\sim\pi(.|X_t)$, $\forall t\geq0, R_t=r(X_t, A_t)$ and $\forall t\geq0, X_{t+1}\sim P(.|X_t, A_t)$. Then, the state-action value function $Q_r^{\pi}(x, a)$ for the policy $\pi$ and the  state-action tuple $(x, a)$ is defined as:
\begin{equation*}
    Q_r^{\pi}(x, a) = \mathbb{E}_{\tau\sim\T(x, a, \pi)}\left[ \sum_{t\geq 0} R_t\right].
\end{equation*}
The optimal state-action value function $Q^*$ is defined as:
\begin{equation*}
    Q_r^*(x, a) = \max_{\pi}Q_r^{\pi}(x, a).
\end{equation*}
where the $\max$ is taken over all stochastic policies.

Let define the one-step evaluation Bellman operator $T^\pi_r$, for all functions $Q\in\mathbb{R}^{\mathcal{X}\times\mathcal{A}}$ and for all state-action tuples $(x, a)\in \mathcal{X}\times\mathcal{A}$, as:
\begin{equation*}
    T^\pi_r Q(x, a) = r(x, a) + \gamma \sum_{b\in\mathcal{A}}\sum_{x'\in\mathcal{X}}\pi(b|x)P(x'|x, a)Q(x', b).
\end{equation*}
The one-step evaluation Bellman operator can also be written with vectorial notations:
\begin{equation*}
    T^\pi_r Q = r + \gamma P^\pi Q,
\end{equation*}
where $P^\pi$ is a transition matrix representing the effect of acting according to $\pi$ in a MDP with dynamics $P$. The evaluation Bellman operator is a contraction and its fixed point is $Q^{\pi}_r$. 

Finally let define the greedy operator $\mathcal{G}$, for all functions $Q\in\mathbb{R}^{\mathcal{X}\times\mathcal{A}}$ and for all state $x\in \mathcal{X}$, as:
\begin{equation*}
     \mathcal{G}(Q)(x) = \argmax_{a\in\mathcal{A}}Q(x,a).
\end{equation*}
Then, one can show~\citep{puterman1990markov}, via a fixed point argument, that the following discrete scheme:
$$
\forall k\geq0, \quad\left\{
    \begin{array}{ll}
        \pi_k = \mathcal{G}\left(Q_k\right), &\\
         Q_{k+1} = T_{r}^{\pi_k}Q_k,&
    \end{array}
\right.
$$
where $Q_0$ can be initialized arbitrarily, converges to $Q^*_r$. This discrete scheme is called the one-step value iteration scheme.

Throughout the article, we also use transformed Bellman operators (see Sec.~\ref{subsec:lossfunction}). The one-step transformed evaluation Bellman operator $T^\pi_{r, h}$, for all functions $Q\in\mathbb{R}^{\mathcal{X}\times\mathcal{A}}$ and for all state-action tuples $(x, a)\in \mathcal{X}\times\mathcal{A}$, can be defined as:
\begin{equation*}
    T^\pi_{r, h} Q(x, a) = h\left(r(x, a) + \gamma \sum_{b\in\mathcal{A}}\sum_{x'\in\mathcal{X}}\pi(b|x)P(x'|x, a)h^{-1}(Q(x', b))\right),
\end{equation*}
where $h$ is a monotonically increasing and invertible  squashing function that scales the state-action value function to make it easier to approximate for a neural network.  
In particular, we use the function $h$:
\begin{align*}
\forall z\in\mathbb{R},\quad h(z)&=\sign(z)(\sqrt{|z|+1}-1) + \epsilon z,
\\
\forall z\in\mathbb{R},\quad h^{-1}(z)&=\sign(z)\left(\left(\frac{\sqrt{1+4\epsilon(|z|+1+\epsilon)}-1}{2\epsilon}\right)-1\right),
\end{align*}
with $\epsilon$ a small number.
The one-step transformed evaluation Bellman operator can also be written with vectorial notations:
\begin{equation*}
    T^\pi_{r, h} Q = h\left(r + \gamma P^\pi h^{-1}(Q)\right).
\end{equation*}

Under some conditions on $h$~\citep{pohlen2018observe} and via a contraction argument, one can show that the transformed one-step value iteration scheme:
$$
\forall k\geq0, \quad\left\{
    \begin{array}{ll}
        \pi_k = \mathcal{G}\left(Q_k\right), &\\
         Q_{k+1} = T_{r, h}^{\pi_k}Q_k,&
    \end{array}
\right.
$$
where $Q_0$ can be initialized arbitrarily, converges. We note this limit $Q^*_{r, h}$.
\section{Extrinsic-Intrinsic Decomposition}
\label{app:decoupling}
For an intrinsically-motivated agent, the reward function $r$ is a linear combination of the intrinsic reward $r^i$ and the extrinsic reward $r^e$:
\begin{equation*}
    r = r^e + \beta r^i.
\end{equation*}
One can compute the optimal state-action value function $Q^*_r$ via the value iteration scheme:
$$
\forall k\geq0, \quad\left\{
    \begin{array}{ll}
        \pi_k = \mathcal{G}\left(Q_k\right), &\\
         Q_{k+1} = T_{r}^{\pi_k}Q_k,&
    \end{array}
\right.
$$
where $Q_0$ can be initialized arbitrarily.

Now, we want to show how we can also converge to $Q^*_{r}$ using separate intrinsic and extrinsic state-action value functions. Indeed, let us consider the following discrete scheme:
$$
\forall k\geq0, \quad\left\{
    \begin{array}{ll}
        \tilde{\pi}_k = \mathcal{G}\left(Q^e_k + \beta Q^i_k \right), &\\
         Q^i_{k+1} = T_{r^i}^{\tilde{\pi}_k}Q^i_k,&\\
         Q^e_{k+1} = T_{r^e}^{\tilde{\pi}_k}Q^e_k,&
    \end{array}
\right.
$$
where the functions $(Q^e_0, Q^i_0)$ can be initialized arbitrarily.

Our goal is simply to show that the linear combination of extrinsic and intrinsic state-action value function $\tilde{Q}_{k}$:
\begin{equation*}
\forall k\geq0, \tilde{Q}_k = Q^e_k + \beta Q^i_k.
\end{equation*}
verifies a one-step value iteration scheme with respect to the reward $r= r^e+\beta r^i$ and therefore converges to $Q^*_r$. To show that let us rewrite $\tilde{Q}_{k+1}$:
\begin{align*}
\tilde{Q}_{k+1} &= Q^e_{k+1} + \beta Q^i_{k+1},
\\
&=T_{r^e}^{\tilde{\pi}_k}Q^e_k + \beta T_{r^i}^{\tilde{\pi}_k}Q^i_k,
\\
& = r^e + \beta r^i + \gamma P^{\tilde{\pi}_k} (Q^e_k + \beta Q^i_k),
\\
& = T_{r^e + \beta r^i}^{\tilde{\pi}_k}(Q^e_k + \beta Q^i_k),
\\
& = T_r^{\tilde{\pi}_k}\tilde{Q}_{k}.
\end{align*}
Therefore we have that $\tilde{Q}_{k}$ satisfies a value iteration scheme with respect to the reward $r= r^e+\beta r^i$:
$$
\forall k\geq0, \quad\left\{
    \begin{array}{ll}
        \tilde{\pi}_k = \mathcal{G}\left(\tilde{Q}_k\right), &\\
         \tilde{Q}_{k+1} = T_{r}^{\tilde{\pi}_k}\tilde{Q}_k,&
    \end{array}
\right.
$$
and by the contraction property:
\begin{equation*}
    \lim_{k\longrightarrow\infty}\tilde{Q}_k = Q^*_r.
\end{equation*}

This result means that we can compute separately $Q^e_k$ and $Q^i_k$ and then mix them to obtain the same behavior than if we had computed $Q_k$ directly with the mixed reward $r^e + \beta r^i$. This implies that we can separately compute the extrinsic and intrinsic component. Each architecture will need to learn their state-action value for different mixtures $\beta$ and then act according to the greedy policy of the mixture of the state-action value functions. This result could also be thought as related to~\citet{barreto2017successor} which may suggest potential future research directions.

The same type of result holds for the transformed state-action value functions. Indeed let us consider the optimal transformed state-action value function $Q^*_{r, h}$ that can be computed via the following discrete scheme:
$$
\forall k\geq0, \quad\left\{
    \begin{array}{ll}
        \pi_k = \mathcal{G}\left(Q_k\right), &\\
         Q_{k+1} = T_{r, h}^{\pi_k}Q_k,&
    \end{array}
\right.
$$
where $Q_0$ can be initialized arbitrarily.

Now, we show how we can compute $Q^*_{r, h}$ differently using separate intrinsic and extrinsic state-action value functions. Indeed, let us consider the following discrete scheme:
$$
\forall k\geq0, \quad\left\{
    \begin{array}{ll}
        \tilde{\pi}_k = \mathcal{G}\left(h\left(h^{-1}(Q^e_k) + \beta h^{-1}(Q^i_k)\right) \right), &\\
         Q^i_{k+1} = T_{r^i, h}^{\tilde{\pi}_k}Q^i_k,&\\
         Q^e_{k+1} = T_{r^e, h}^{\tilde{\pi}_k}Q^e_k,&
    \end{array}
\right.
$$
where the functions $(Q^e_0, Q^i_0)$ can be initialized arbitrarily.

We want to show that $\tilde{Q}_{k}$ defines as:
\begin{equation*}
\forall k\geq0,\quad  \tilde{Q}_{k} = h\left(h^{-1}(Q^e_k) + \beta h^{-1}(Q^i_k)\right),
\end{equation*}
verifies the one-step transformed value iteration scheme with respect to the reward $r= r^e+\beta r^i$ and therefore converges to $Q^*_{r, h}$. To show that let us rewrite $\tilde{Q}_{k+1}$:
\begin{align*}
\tilde{Q}_{k+1} &= h\left(h^{-1}(Q^e_{k+1}) + \beta h^{-1}(Q^i_{k+1})\right),
\\
&=h\left(h^{-1}(T_{r^e, h}^{\tilde{\pi}_k}Q^e_k) + \beta h^{-1}(T_{r^i, h}^{\tilde{\pi}_k}Q^i_k)\right),
\\
& = h\left(r^e + \gamma P^{\tilde{\pi}_k} h^{-1}(Q^e_k) + \beta r^i + \gamma P^{\tilde{\pi}_k} \beta h^{-1}(Q^i_k)\right),
\\
& = h\left(r^e + \beta r^i + \gamma P^{\tilde{\pi}_k}(h^{-1}(Q^e_k)+ \beta h^{-1}(Q^i_k))\right),
\\
& = h\left(r + \gamma P^{\tilde{\pi}_k} h^{-1}(\tilde{Q}_k)\right)
\\
& = T_{r, h}^{\tilde{\pi}_k}\tilde{Q}_{k}.
\end{align*}

Thus we have that $\tilde{Q}_{k}$ satisfies the one-step transformed value iteration scheme with respect to the reward $r= r^e+\beta r^i$:
$$
\forall k\geq0, \quad\left\{
    \begin{array}{ll}
        \tilde{\pi}_k = \mathcal{G}\left(\tilde{Q}_k\right), &\\
         \tilde{Q}_{k+1} = T_{r, h}^{\tilde{\pi}_k}Q_k,&
    \end{array}
\right.
$$
and by contraction:
\begin{equation*}
    \lim_{k\longrightarrow\infty}\tilde{Q}_k = Q^*_{r, h}.
\end{equation*}

One can remark that when the transformation $h$ is the identity, we recover the linear mix between intrinsic and extrinsic state-action value functions.

\section{Retrace and Transformed Retrace}
\label{app:retrace}
Retrace~\citep{munos2016safe} is an off-policy RL algorithm for evaluation or control. In the evaluation setting the goal is to estimate the state-action value function $Q^\pi$ of a target policy $\pi$ from trajectories drawn from a behaviour policy $\mu$. In the control setting the goal is to build a sequence of target policies $\pi_k$ and state-action value functions $Q_k$ in order to approximate $Q^*$.

The evaluation Retrace operator $T^{\mu,\pi}_r$, that depends on $\mu$ and $\pi$, is defined as follows, for all functions $Q\in\mathbb{R}^{\mathcal{X}\times\mathcal{A}}$ and for all state-action tuples $(x, a)\in \mathcal{X}\times\mathcal{A}$:
\begin{equation*}
T^{\mu,\pi}_rQ(x,a) = \mathbb{E}_{\tau\sim\T(x, a, \mu)}\left[Q(x,a) +\sum_{t\geq0}\gamma^t\left(\prod_{s=1}^tc_s\right)\delta_t \right],
\end{equation*}
where the temporal difference $\delta_t$ is defined as:
\begin{equation*}
\delta_t = r_t + \gamma \sum_{a\in A}\pi(a|X_{t+1})Q(X_{t+1},a)-Q(X_t, A_t),    
\end{equation*}
and the trace coefficients $c_s$ as:
\begin{equation*}
c_s = \lambda\min\left(1, \frac{\pi(A_s|X_s)}{\mu(A_s|X_s)}\right), 
\end{equation*}
where $\lambda$ is a fixed parameter $\in [0, 1]$.
The operator $T^{\mu,\pi}_r$ is a multi-step evaluation operator that corrects the behaviour of $\mu$ to evaluate the policy $\pi$. It has been shown in Theorem 1 of~\citet{munos2016safe} that $Q^\pi_r$ is the fixed point of $T^{\mu,\pi}_r$. In addition, Theorem 2 of~\citet{munos2016safe} explains in which conditions the Retrace value iteration scheme:
$$
\forall k\geq0, \quad\left\{
    \begin{array}{ll}
        \pi_k = \mathcal{G}\left(Q_k\right), &\\
         Q_{k+1} = T^{\mu_k,\pi_k}_r Q_k,&
    \end{array}
\right.
$$
converges to the optimal state-action value function $Q^*$, where $Q_0$ is initialized arbitrarily and $\{\mu_k\}_{k\in\mathbb{N}}$ is an arbitrary sequence of policies that may depend on $Q_k$.

As in the case of the one-step Bellman operator, we can also define a transformed counterpart to the Retrace operator. More specifically, we can define the transformed Retrace operator $T^{\mu,\pi}_{r, h}$, for all functions $Q\in\mathbb{R}^{\mathcal{X}\times\mathcal{A}}$ and for all state-action tuples $(x, a)\in \mathcal{X}\times\mathcal{A}$:
\begin{equation*}
T^{\mu,\pi}_{r, h}Q(x,a) =  h\left(\mathbb{E}_{\tau\sim\T(x, a, \mu)}\left[h^{-1}(Q(x,a)) +\sum_{t\geq0}\gamma^t\left(\prod_{s=1}^tc_s\right)\delta^h_t\right]\right),
\end{equation*}
where the temporal difference $\delta^h_t$ is defined as:
\begin{equation*}
\delta^h_t = r_t + \gamma \sum_{a\in A}\pi(a|X_{t+1})h^{-1}(Q(X_{t+1},a))-h^{-1}(Q(X_t, A_t)).  
\end{equation*}
As in the case of the Retrace operator, we can define the transformed Retrace value iteration scheme:
$$
\forall k\geq0, \quad\left\{
    \begin{array}{ll}
        \pi_k = \mathcal{G}\left(Q_k\right), &\\
         Q_{k+1} = T^{\mu_k,\pi_k}_{r, h} Q_k,&
    \end{array}
\right.
$$
where $Q_0$ is initialized arbitrarily and $\{\mu_k\}_{k\in\mathbb{N}}$ is an arbitrary sequence of policies.

\subsection{Extrinsic-Intrinsic Decomposition for Retrace and Transformed Retrace}
\label{subsec:decompositionretrace}
Following the same methodology than App~.\ref{app:decoupling}, we can also show that the state-action value function can be decomposed in extrinsic and intrinsic components for the Retrace and transformed Retrace value iteration schemes when the reward is of the form $r=r^e+\beta r^i$. 

Indeed if we define the following discrete scheme:
$$
\forall k\geq0, \quad\left\{
    \begin{array}{ll}
        \tilde{\pi}_k = \mathcal{G}\left(Q^e_k + \beta Q^i_k \right), &\\
         Q^i_{k+1} = T_{r^i}^{\tilde{\mu}_k, \tilde{\pi}_k}Q^i_k,&\\
         Q^e_{k+1} = T_{r^e}^{\tilde{\mu}_k, \tilde{\pi}_k}Q^e_k,&
    \end{array}
\right.
$$
where the functions $(Q^e_0, Q^i_0)$ can be initialized arbitrarily and $\{\tilde{\mu_k}\}_{k\in\mathbb{N}}$ is an arbitrary sequence of policies. Then, it is straightforward to show that the linear combination $\tilde{Q}_k$:
\begin{equation*}
\forall k\geq0, \tilde{Q}_k = Q^e_k + \beta Q^i_k,
\end{equation*}
verifies the Retrace value iteration scheme:
$$
\forall k\geq0, \quad\left\{
    \begin{array}{ll}
        \tilde{\pi}_k  = \mathcal{G}\left(\tilde{Q}_k\right), &\\
         \tilde{Q}_{k+1} = T^{\tilde{\mu}_k, \tilde{\pi}_k}_r \tilde{Q}_k,&
    \end{array}
\right.
$$

Likewise, if we define the following discrete scheme:
$$
\forall k\geq0, \quad\left\{
    \begin{array}{ll}
        \tilde{\pi}_k = \mathcal{G}\left(h\left(h^{-1}(Q^e_k) + \beta h^{-1}(Q^i_k)\right) \right), &\\
         Q^i_{k+1} = T_{r^i, h}^{\tilde{\mu}_k, \tilde{\pi}_k}Q^i_k,&\\
         Q^e_{k+1} = T_{r^e, h}^{\tilde{\mu}_k, \tilde{\pi}_k}Q^e_k,&
    \end{array}
\right.
$$
where the functions $(Q^e_0, Q^i_0)$ can be initialized arbitrarily and $\{\tilde{\mu_k}\}_{k\in\mathbb{N}}$ is an arbitrary sequence of policies. Then, it is also straightforward to show that $\tilde{Q}_{k}$ defines as:
\begin{equation*}
\forall k\geq0,\quad  \tilde{Q}_{k} = h\left(h^{-1}(Q^e_k) + \beta h^{-1}(Q^i_k)\right),
\end{equation*}
verifies the transformed Retrace value iteration scheme:
$$
\forall k\geq0, \quad\left\{
    \begin{array}{ll}
        \tilde{\pi}_k = \mathcal{G}\left(\tilde{Q}_k\right), &\\
         \tilde{Q}_{k+1} = T^{\tilde{\mu}_k, \tilde{\pi}_k}_{r, h}Q_k,&
    \end{array}
\right.
$$

\subsection{Retrace and Transformed Retrace Losses for Neural Nets.}
\label{subsec:lossfunction}

In this section, we explain how we approximate with finite data and neural networks the Retrace value iteration scheme. To start, one important thing to remark is that we can rewrite the evaluation step:
\begin{equation*}
    Q_{k+1} = T^{\mu_k, \pi_k}_r Q_k,
\end{equation*}
with:
\begin{equation*}
    \label{eq:optimal}
    Q_{k+1} = \argmin_{Q\in\mathbb{R}^{\mathcal{X}\times\mathcal{A}}} \|T^{\mu_k, \pi_k}_r Q_k-Q\|,
\end{equation*}
where $\|.\|$ can be any norm over the function space $\mathbb{R}^{\mathcal{X}\times\mathcal{A}}$. This means that the evaluation step can be seen as an optimization problem over a functional space where the optimization consists in finding a function $Q$ that matches the target $T^{\mu_k, \pi_k}_r Q_k$.

In practice, we face two important problems. The search space $\mathbb{R}^{\mathcal{X}\times\mathcal{A}}$ is too big and we cannot evaluate $T^{\mu_k, \pi_k}_r Q_k$ everywhere because we have a finite set of data. To tackle the former, a possible solution is to use function approximation such as neural networks. Thus, we parameterize the state action value function $Q(x,a; \theta)$ (where $\theta$ is the set of parameters of the neural network) also called online network. Concerning the latter, we are going to build sampled estimates of $T^{\mu_k, \pi_k}_r Q_k$ and use them as targets for our optimization problem. In practice, the targets are built from a previous and fixed set of parameters $\theta^-$ of the neural network. $Q(x,a;\theta^-)$ is called the target network. The target network is updated to the value of the online network at a fixed frequency during the learning.

More precisely, let us consider a batch of size $B$ of finite sampled sequences of size $H$: $D=\{(x^b_s, a^b_s, \mu^b_s=\mu(a^b_s|x^b_s), r^b_s, x^b_{s+1})_{s=t}^{t+H-1}\}_{b=0}^{B-1}$ starting from $(x^b_t, a^b_t)$ and then following the behaviour policy $\mu$.
Then, we can define the finite sampled-Retrace targets as:
\begin{align*}
\hat{T}^{\mu, \pi}_r Q(x^b_s, a^b_s; \theta^-)&= Q(x^b_s, a^b_s; \theta^-) + \sum_{j=s}^{t+H-1}\gamma^{j-s}\left(\prod_{i=s+1}^j c_{i,b}\right)\delta_{j, b}
\\
c_{i, b} &= \lambda\min\left(1, \frac{\pi(a^b_i|x^b_i)}{\mu^b_i}\right), 
\\
\delta_{j, b}&= r^b_j + \gamma \sum_{a\in A}\pi(a|x^b_{j+1})Q(x^b_{j+1},a; \theta^-)-Q(x^b_j, a^b_j; \theta^-),
\end{align*}
where $\pi(a|x)$ is the target policy.

Once the targets are computed, the goal is to find a parameter $\theta$ that fits those targets by minimizing the following loss function:
\begin{equation*}
    L(D, \theta, \theta^-, \pi, \mu, r) = \sum_{b=0}^{B-1}\sum_{s=t}^{t+H-1}\left(Q(x^b_s, a^b_s;\theta)-\hat{T}^{\mu, \pi}_r Q(x^b_s, a^b_s; \theta^-)\right)^2.
\end{equation*}
This is done by an optimizer such as gradient descent for instance. Once $\theta$ is updated by the optimizer, a new loss with new targets is computed and minimized until convergence.

Therefore in practice the evaluation step of the Retrace value iteration scheme $Q_{k+1} = T^{\mu_k, \pi_k}_r Q_k$ is approximated by minimizing the loss $L(D, \theta, \pi, \mu)$ with an optimizer. The greedy step $\pi_k = \mathcal{G}\left(Q_k\right)$ is realized by simply being greedy with respect to the online network and choosing the target policy as follows: $\pi = \mathcal{G}\left(Q(x,a;\theta)\right)$.

In the case of a transformed Retrace operator, we have  the following targets:
\begin{align*}
\hat{T}^{\mu, \pi}_{r, h} Q(x^b_s, a^b_s; \theta^-)&=h\left( h^{-1}(Q(x^b_s, a^b_s; \theta^-)) + \sum_{j=s}^{t+H-1}\gamma^{j-t}\left(\prod_{i=s+1}^j c_{i,b}\right)\delta^h_{s, b}\right)
\\
c_{i, b} &= \lambda\min\left(1, \frac{\pi(a^b_i|x^b_i)}{\mu^b_i}\right), 
\\
\delta_{j, b}&= r^b_j + \gamma \sum_{a\in A}\pi(a|x^b_{j+1})h^{-1}(Q(x^b_{j+1},a; \theta^-))-h^{-1}Q(x^b_j, a^b_j; \theta^-).
\end{align*}

And the transformed Retrace loss function is:
\begin{equation*}
    L(D, \theta, \theta^- \pi, \mu, r, h) = \sum_{b=0}^{B-1}\sum_{s=t}^{t+H-1}\left(Q(x^b_s,a^b_s;\theta)-\hat{T}^{\mu, \pi}_{r, h} Q(x^b_s, a^b_s; \theta^-)\right)^2.
\end{equation*}
\section{Multi-arm Bandit Formalism}
\label{app:bandits}
This section describes succinctly the multi-arm bandit (MAB) paradigm, upper confidence bound (UCB) algorithm and sliding-window UCB algorithm. For a more thorough explanation and analysis we refer the reader to~\citet{garivier2008upperconfidence}. 

At each time $k\in\mathbb{N}$, a MAB algorithm chooses an arm $A_k$ among the possible arms $\{0,\dots, N-1\}$ according to a policy $\pi$ that is conditioned on the sequence of previous actions and rewards. Doing so, it receives a reward  $R_k(A_k)\in\mathbb{R}$. In the stationary case, the rewards $\{R_k(a)\}_{k\geq0}$ for a given arm $a\in\{0,\dots, N-1\}$ are modelled by a sequence of i.i.d random variables. In the non-stationary case, the rewards $\{R_k(a)\}_{k\geq0}$ are modelled by a sequence of independent random variables but whose distributions could change through time.

The goal of a MAB algorithm is to find a policy $\pi$ that maximizes the expected cumulative reward for a given horizon $K$:
\begin{equation*}
    \mathbb{E}_\pi\left[\sum_{k=0}^{K-1} R_k(A_k)\right].
\end{equation*}

In the stationary case, the UCB algorithm has been well studied and is commonly used. Let us define the number of times an arm $a$ has been played after $k$ steps:
\begin{equation*}
N_k(a) = \sum_{m=0}^{k-1}\mathbf{1}_{\{A_m=a\}}.
\end{equation*}
Let us also define the empirical mean of an arm $a$ after $k$ steps:
\begin{equation*}
\hat{\mu}_k(a) = \frac{1}{N_k(a)}  \sum_{m=0}^{k-1} R_k(a)\mathbf{1}_{\{A_m=a\}}.  
\end{equation*}

The UCB algorithm is then defined as follows:
$$
\left\{
    \begin{array}{ll}
         \forall 0\leq k\leq N-1,\quad A_k=k &\\
         \forall N\leq k\leq K-1,\quad A_k =\argmax_{1\leq a \leq N} \hat{\mu}_{k-1}(a) + \beta\sqrt{\frac{\log{(k-1)}}{N_{k-1}(a)}} &
    \end{array}
\right.
$$

In the non-stationary case, the UCB algorithm cannot adapt to the change of reward distribution and one can use a sliding-window UCB in that case. It is commonly understood that the window length $\tau\in\mathbb{N}^*$ should be way smaller that the horizon $K$.
Let us define the number of times an arm $a$ has been played after $k$ steps for a window of length $\tau$:
\begin{equation*}
N_k(a, \tau) = \sum_{m=0\vee k-\tau}^{k-1}\mathbf{1}_{\{A_m=a\}},
\end{equation*}
where $0\vee k-\tau$ means $\max(0,k-\tau)$. Let define the empirical mean of an arm $a$ after $k$ steps for a window of length $\tau$:
\begin{equation*}
\hat{\mu}_k(a, \tau) = \frac{1}{N_k(a, \tau)}  \sum_{m=0\vee k-\tau}^{k-1} R_k(a)\mathbf{1}_{\{A_m=a\}}.  
\end{equation*}

Then , the sliding window UCB can be defined as follows:
$$
\left\{
    \begin{array}{ll}
         \forall 0\leq k\leq N-1,\quad A_k=k &\\
         \forall N\leq k\leq K-1,\quad A_k =\argmax_{1\leq a \leq N} \hat{\mu}_{k-1}(a, \tau) + \beta\sqrt{\frac{\log{(k-1\wedge\tau)}}{N_{k-1}(a, \tau)}} &
    \end{array}
\right.
$$
where $k-1\wedge\tau$ means $\min(k-1,\tau)$.

In our experiments, we use a simplified sliding window UCB with $\epsilon_{\texttt{UCB}}$-greedy exploration:
$$
\left\{
    \begin{array}{ll}
         \forall 0\leq k\leq N-1,\quad A_k=k &\\
         \forall N\leq k\leq K-1 \text{ and } U_k\geq \epsilon_{\texttt{UCB}},\quad A_k =\argmax_{0\leq a \leq N-1} \hat{\mu}_{k-1}(a, \tau) + \beta\sqrt{\frac{1}{N_{k-1}(a, \tau)}} &\\
         \forall N\leq k\leq K-1 \text{ and } U_k< \epsilon_{\texttt{UCB}},\quad A_k = Y_k &
    \end{array}
\right.
$$
where $U_k$ is a random value drawn uniformly from $[0, 1]$ and $Y_k$ a random action drawn uniformly from $\{0, \dots, N-1\}$.
\section{Implementation details of the distributed setting}
\label{app:distributed}

{\bf Replay buffer:} it stores fixed-length sequences of \emph{transitions} $\xi=(\omega_s)_{s=t}^{t+H-1}$ along with their priorities $p_\xi$.
A transition is of the form $\omega_s=(r^e_{s-1}, r^i_{s-1}, a_{s-1}, h_{s-1}, x_s, a_s, h_s, \mu_s, j_s, r^e_{s}, r^i_{s}, x_{s+1})$ . Such transitions are also called \emph{timesteps} and the length of a sequence $H$ is called the \emph{trace length}. In addition, adjacent sequences in the replay buffer overlap by a number of timesteps called the \emph{replay period} and the sequences never cross episode boundaries. Let us describe each element of a transition:
\begin{itemize}
    \item $r^e_{s-1}$: extrinsic reward at the previous time.
    \item $r^i_{s-1}$: intrinsic reward at the previous time.
    \item $a_{s-1}$: action done by the agent at the previous time.
    \item $h_{s-1}$: recurrent state (in our case hidden state of the LSTM) at the previous time.
    \item $x_s$: observation provided by the environment at the current time.
    \item $a_s$: action done by the agent at the current time.
    \item $h_{s}$: recurrent state (in our case hidden state of the LSTM) at the current time.
    \item $\mu_s$: the probability of choosing the action $a_s$.
    \item $j_s=j$: index of the pair $(\gamma_{j}, \beta_{j})$ chosen at a beginning of an episode in each actor by the multi-arm bandit algorithm (fixed for the whole sequence).
    \item $r^e_s$: extrinsic reward at the current time.
    \item $r^i_s$: intrinsic reward at the current time
    \item $x_{s+1}$: observation provided by the environment at the next time.
\end{itemize}
In our experiment, we choose a trace length of $160$ with a replay period of $80$ or a trace length of $80$ with a replay period of $40$.
Please refer to~\citep{kapturowski2018recurrent} for a detailed experimental of trade-offs on different treatments of recurrent states in the replay. Finally, concerning the priorities, we followed the same prioritization scheme proposed by~\citet{kapturowski2018recurrent} using a mixture of max and mean of the TD-errors in the sequence with priority exponent $\eta= 0.9$.

{\bf Actors}: each of the $L$ actors shares the same network architecture as the learner but with different weights $\theta_l$, with $0 \leq l \leq L-1$. The $l$-th actor updates its weights $\theta_l$ every 400 frames by copying the weights of the learner. At the beginning of each episode, each actor chooses, via a multi-arm bandit algorithm, an index $j$ that represents a pair $(\gamma_j, \beta_j)$ in the family of pairs $(\{\beta_j, \gamma_j)\}_{j=0}^{N-1}$. In addition, the recurrent state is initialized to zero.
To act, an actor will need to do a forward pass on the network in order to compute the state-action value for all actions, noted $Q(x_t, ., j; \theta_l)$. To do so the inputs of the network are :
\begin{itemize}
    \item $x_t$: the observation at time $t$.
    \item $r^e_{t-1}$: the extrinsic reward at the previous time, initialized with $r^e_{-1}=0$.
    \item $r^i_{t-1}$: the intrinsic reward at the previous time, initialized with $r^i_{-1}=0$.
    \item $a_{t-1}$: the action at the previous time, $a_{-1}$ is initialized randomly.
    \item $h_{t-1}$: recurrent state at the previous time, is initialized with $h_{-1}=0$.
    \item $j_{t-1}=j$: the index of the pair $(\beta_j, \gamma_j)$ chosen by the multi-arm bandit algorithm (fixed for all the episode).
\end{itemize}

At time $t$, the $l-$th actor acts $\epsilon_l$-greedy with respect to $Q(x_t, ., j; \theta_l)$:
$$
\left\{
    \begin{array}{ll}
         \text{If: } U_t< \epsilon_l, a_t = Y_t, &\\
         \text{Else: } a_t = \argmax_{a\in\mathcal{A}} Q(x_t, a, j; \theta_l),&
    \end{array}
\right.
$$
where $U_t$ is a random value drawn uniformly from $[0, 1]$ and $Y_t$ a random action drawn uniformly from $\mathcal{A}$. The probability $\mu_t$ associated to $a_t$ is therefore:
$$
\left\{
    \begin{array}{ll}
         \text{If: } U_t< \epsilon_l, \mu_t = \frac{\epsilon_l}{|\mathcal{A}|}, &\\
         \text{Else: } \mu_t = 1 - \epsilon_l\frac{|\mathcal{A}|-1}{|\mathcal{A}|},&
    \end{array}
\right.
$$
where $|\mathcal{A}|$ is the cardinal number of the action space, $18$ in the case of Atari games.
Then, the actor plays the action $a_t$ and computes the intrinsic reward $r^i_t$ and the environment produces the next observation $x_{t+1}$ and the extrinsic reward $r^e_t$. This process goes on until the end of the episode.

The value of the noise $\epsilon_l$ is chosen according to the same formula established by~\citet{horgan2018distributed}:
\begin{equation*}
   \epsilon_l = \epsilon^{1+\alpha\frac{l}{L-1}} 
\end{equation*}
where $\epsilon=0.4$ and  $\alpha=8$. In our experiments, we fix the number of actors to $L=256$. Finally, the actors send the data collected to the replay along with the priorities.

{\bf Evaluator}: the evaluator shares the same network architecture as the learner but with different weights $\theta_e$. The evaluator updates its weights $\theta_l$ every $5$ episodes frames by copying the weights of the learner. Unlike the actors, the experience produced by the evaluator is not sent to the replay buffer. The evaluator alternates between the following states every $5$ episodes:

\begin{itemize}
    \item \textbf{Training bandit algorithm}: the evaluator chooses, via a multi-arm bandit algorithm, an index $j$ that represents a pair $(\gamma_j, \beta_j)$ in the family of pairs $(\{\beta_j, \gamma_j)\}_{j=0}^{N-1}$. Then it proceeds to act in the same way as the actors, described above. At the end of the episode, the undiscounted returns are used to train the multi-arm bandit algorithm. 
    \item \textbf{Evaluation}: the evaluator chooses the greedy choice of index $j$,  $\argmax_{1\leq a \leq N} \hat{\mu}_{k-1}(a)$, so it acts with $(\gamma_j, \beta_j)$. Then it proceeds to act in the same way as the actors, described above. At the end of $5$ episodes and before switching to the other mode, the results of those $5$ episodes are average and reported.
\end{itemize}

{\bf Learner}: The learner contains two identical networks called the online and target networks with different weights $\theta$ and $\theta^-$ respectively \cite{mnih2015human}.
The target network's weights $\theta^-$ are updated to $\theta$ every $1500$ optimization steps. For our particular architecture, the weights $\theta=\theta^e\cup\theta^i$ can be decomposed in a set of intrinsic weights $\theta^e$ and $\theta^i$ that have the same architecture. Likewise, we have $\theta^-=\theta^{-,e}\cup\theta^{-, i}$. The intrinsic and extrinsic weights are going to be updated by their own transformed Retrace loss. $\theta^e$ and $\theta^i$ are updated by executing the following sequence of instructions:
\begin{itemize}
    \item First, the learner samples a batch of size $B$ of fixed-length sequences of transitions $D=\{\xi^b=(\omega^b_s)_{s=t}^{t+H-1}\}_{b=0}^{B-1}$ from the replay buffer.
    \item Then, a forward pass is done on the online network and the target with inputs $\{(x^b_s, r^{e, b}_{s-1}, r^{i, b}_{s-1}, j^b, a^b_{s-1}, h^b_{s-1})_{s=t}^{t+H}\}_{b=0}^{B-1}$ in order to obtain the state-action values $\{(Q(x^b_s,.,j^b; \theta^e), Q(x^b_s,.,j^b; \theta^{-,e}), Q(x^b_s,.,j^b; \theta^{i}), Q(x^b_s,.,j^b; \theta^{-,i}))_{s=t}^{t+H}\}_{b=0}^{B-1}$.
    \item Once the state-action values are computed, it is now easy to compute the transformed Retrace losses $L(D, \theta^e, \theta^{-, e}, \pi, \mu, r^e, h)$ and $L(D, \theta^i, \theta^{-, i}, \pi, \mu, r^i, h)$ for each set of weights $\theta^e$ and $\theta^i$, respectively, as shown in Sec~.\ref{app:retrace}. The target policy $\pi$ is greedy with respect to $Q(x^b_s,.,j^b; \theta^e) + \beta_{j^b_s}Q(x^b_s,.,j^b; \theta^{i})$ or with respect to $h\left(h^{-1}(Q(x^b_s,.,j^b; \theta^e)) + \beta_{j^b_s}h^{-1}(Q(x^b_s,.,j^b; \theta^{i}))\right)$ in the case where we want to apply a transform $h$ to the mixture of intrinsic and extrinsic state-action value functions.
    \item The transformed Retrace losses are optimized with an Adam optimizer.
    \item Like NGU, the inverse dynamics model and the random network distillation losses necessary to compute the intrinsic rewards are optimized with an Adam optimizer.
    \item Finally, the priorities are computed for each sampled sequence of transitions $\xi^b$ and updated in the replay buffer.
\end{itemize}

\textbf{Computation used}: in terms of hardware we train the agent with a single GPU-based learner, performing approximately $5$ network updates per second (each update on a mini-batch of $64$ sequences of length $160$. We use $256$ actors, with each one performing $\sim 260$ environment steps per second on Atari.
\clearpage
\section{Network Architectures}
\label{app:neural}

\begin{figure}[!ht]
    \centering
    \includegraphics[width=0.7\textwidth]{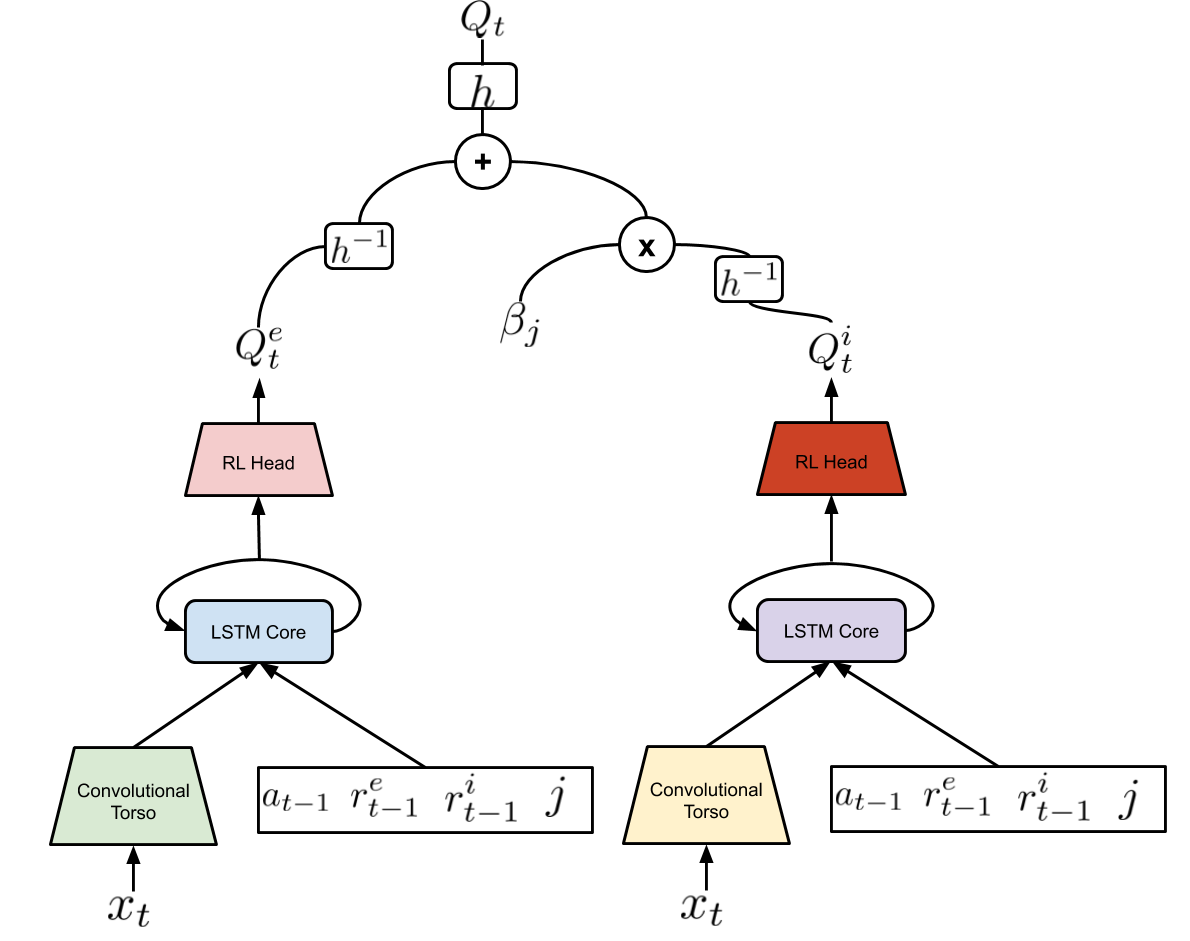}
    \caption{Sketch of the Agent57.} 
    \label{fig:spectrumarchitecture.}
\end{figure}

\begin{figure}[!ht]
    \centering
    \includegraphics[width=0.7\textwidth]{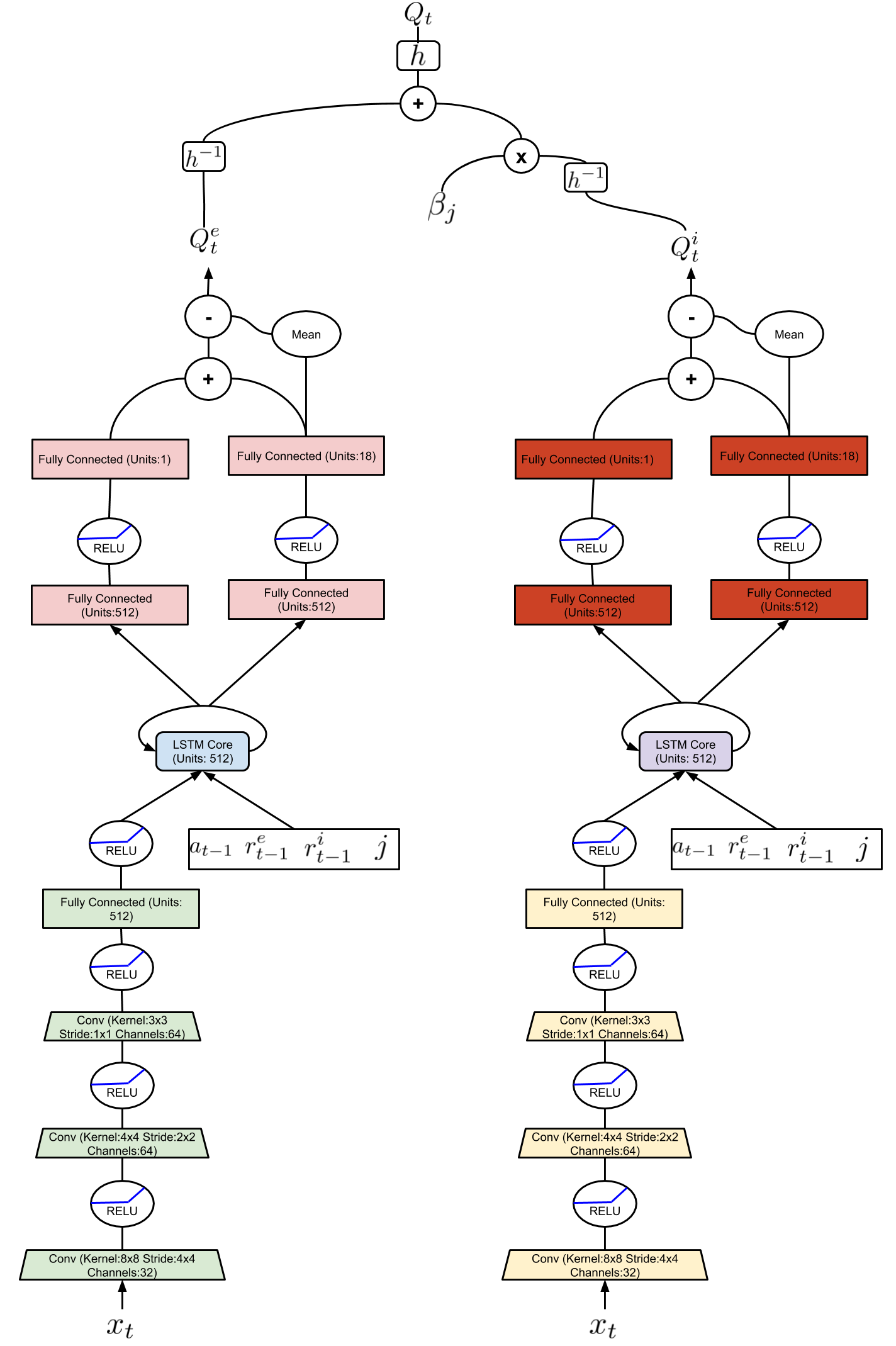}
    \caption{Detailed Agent57.} 
    \label{fig:detailedspectrumarchitecture.}
\end{figure}
\clearpage
\section{Hyperparameters}

\subsection{Values of $\beta$ and $\gamma$}
\label{app:family}
\begin{figure}[h]
\label{app:graphs}
    \centering
    \subfigure[Values taken by the $\{\beta_i\}_{i=0}^{N-1}$]{\includegraphics[width=0.4\textwidth]{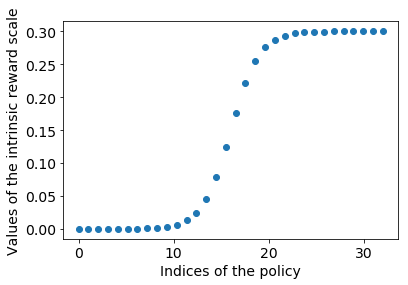}\label{fig:beta_distrib}}
    \subfigure[Values taken by the $\{\gamma_i\}_{i=0}^{N-1}$]{\includegraphics[width=0.4\textwidth]{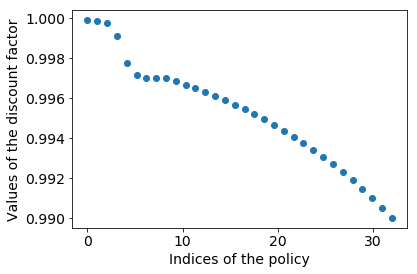}\label{fig:gamma_distrib}}
    \caption{Values taken by the $\{\beta_i\}_{i=0}^{N-1}$ and the $\{\gamma_i\}_{i=0}^{N-1}$ for $N=32$ and $\beta=0.3$.} 
\end{figure}

The intuition between the choice of the set $\{(\beta_j, \gamma_j)\}_{j=0}^{N-1}$ is the following. Concerning the $\beta_j$ we want to encourage policies which are very exploitative and very exploratory and that is why we choose a sigmoid as shown in Fig.~\ref{fig:beta_distrib}. Concerning the $\gamma_j$ we would like to allow for long term horizons (high values of $\gamma_j$) for exploitative policies (small values of $\beta_j$) and small term horizons (low values of $\gamma_j$) for exploratory policies (high values of $\beta_j$). This is mainly due to the sparseness of the extrinsic reward and the dense nature of the intrinsic reward. This motivates the choice done in Fig.~\ref{fig:gamma_distrib}.   

$\beta_j =
\left\{
	\begin{array}{ll}
		0  & \mbox{if }\ j = 0 \\
		\beta=0.3  & \mbox{if }\ j = N-1 \\
		\beta \cdot \sigma(10\frac{2j - (N-2)}{N-2}) & otherwise \\
	\end{array}
\right.$
, $\quad\gamma_j =
\left\{
	\begin{array}{ll}
		\gamma_0  & \mbox{if }\ j = 0 \\
		\gamma_1 + (\gamma_0-\gamma_1)\sigma(10\frac{2i - 6}{6}) & \mbox{if }\ j\in\{1, \dots, 6\} \\
		\gamma_1  & \mbox{if }\ j = 7 \\
		1 - \exp\bigg(\frac{(N-9)\log(1-\gamma_{1}) + (j-8)\log(1-\gamma_{2})}{N-9}\bigg) & otherwise \\
	\end{array}
\right.$

where $N=32$,  $\gamma_0 = 0.9999$, $\gamma_1 = 0.997$ and $\gamma_2 = 0.99$.

\subsection{Atari pre-processing hyperparameters}
\label{atari_hypers}
In this section we detail the hyperparameters we use to pre-process the environment frames received from the Arcade Learning Environment. On Tab.~\ref{table_hyper_atari} we detail such hyperparameters. ALE is publicly available at \url{https://github.com/mgbellemare/Arcade-Learning-Environment}.

\begin{table}[h]
\centering
\begin{tabular}{l|c}
\textbf{Hyperparameter} & \textbf{Value} \\ \hline
Max episode length & $30\ min$  \\ \hline 
Num. action repeats & $4$ \\ \hline
Num. stacked frames & $1$ \\ \hline
Zero discount on life loss & $false$ \\ \hline
Random noops range & $30$ \\ \hline
Sticky actions & $false$ \\ \hline
Frames max pooled & 3 and 4\\ \hline
Grayscaled/RGB & Grayscaled \\ \hline
Action set & Full \\ \hline
\end{tabular}
\caption{Atari pre-processing hyperparameters.}
\label{table_hyper_atari}
\vspace{-2ex}
\end{table}

\clearpage
\subsection{Hyperparameters Used}
\label{app:hyperparameters}

The hyperparameters that we used in all experiments are exactly like those of NGU. However, for completeness, we detail them below in Tab. \ref{tab:hyperparameters}. We also include the hyperparameters we use for the windowed UCB bandit.

\begin{small}
\begin{longtable}[!ht]{l|c}
\centering
\textbf{Hyperparameter} & \textbf{Value} \\ \hline
Number of mixtures $N$ & $32$ \\ \hline
Optimizer & AdamOptimizer (for all losses) \\ \hline
Learning rate (R2D2) & $0.0001$  \\ \hline
Learning rate (RND and Action prediction) & $0.0005$ \\ \hline
Adam epsilon & $0.0001$  \\ \hline
Adam beta1 & $0.9$  \\ \hline
Adam beta2 & $0.999$  \\ \hline
Adam clip norm & $40$  \\ \hline
Discount $r^i$ & $0.99$ \\ \hline
Discount $r^e$ & $0.997$ \\ \hline
Batch size & $64$ \\ \hline
Trace length & $160$ \\ \hline
Replay period & $80$ \\ \hline
Retrace $\lambda$ & $0.95$ \\ \hline
R2D2 reward transformation & ${\rm sign}(x) \cdot (\sqrt{|x| + 1} - 1) + 0.001 \cdot x$ \\ \hline
Episodic memory capacity & $30000$ \\ \hline
Embeddings memory mode & Ring buffer\\ \hline
Intrinsic reward scale $\beta$ & $0.3$ \\ \hline
Kernel $\epsilon$ & $0.0001$ \\ \hline
Kernel num. neighbors used & $10$ \\ \hline 
Replay capacity & $5e6$ \\ \hline
Replay priority exponent & $0.9$ \\ \hline 
Importance sampling exponent & $0.0$ \\ \hline 
Minimum sequences to start replay & $6250$ \\ \hline 
Actor update period & $100$ \\ \hline
Target Q-network update period & $1500$ \\ \hline
Embeddings target update period & once/episode \\ \hline
Action prediction network L2 weight & $0.00001$ \\ \hline
RND clipping factor $L$ & $5$ \\ \hline
Evaluation $\epsilon$ & $0.01$ \\ \hline
Target $\epsilon$ & $0.01$ \\ \hline
Bandit window size & $90$ \\ \hline
Bandit UCB $\beta$ & $1$ \\ \hline
Bandit $\epsilon$ & $0.5$ \\ \hline
\caption{Agent57 hyperparameters.}
\label{tab:hyperparameters}
\vspace{-2ex}
\end{longtable}
\end{small}

\subsection{Hyperparameters Search Range}
The ranges we used to select the hyperparameters of Agent57 are displayed on Tab.~\ref{table_hyper_ranges}.
\begin{table}[!ht]
\centering
\begin{tabular}{l|c}
\textbf{Hyperparameter} & \textbf{Value} \\ \hline
Bandit window size $\tau$ & $\{160,\ 224,\ 320, 640\}$ \\ \hline
Bandit $\epsilon_{\texttt{UCB}}$ & $\{0.3,\ 0.5,\ 0.7\}$ \\ \hline
\end{tabular}
\caption{Range of hyperparameters sweeps.}
\label{table_hyper_ranges}
\end{table}
\section{Experimental Results}

\subsection{Atari 10: Table of Scores for the Ablations}
\label{app:tabatari10}
\small
\begin{tabular}{|c|c|c|c|c|c|}
\hline
 Games & \tiny{R2D2 (Retrace) long trace}  & \tiny{R2D2 (Retrace) high gamma} & \tiny{NGU sep. nets} & \tiny{NGU Bandit} & \tiny{Agent57 small trace} \\
\hline
 beam rider & 287326.72 $\pm$ 5700.31 & \bf{349971.96 $\pm$ 5595.38} & 151082.57 $\pm$ 8666.19 & 249006.62 $\pm$ 19662.62 & 244491.89 $\pm$ 25348.14 \\
 freeway & \bf{33.91 $\pm$ 0.09} & 32.84 $\pm$ 0.06 & 32.91 $\pm$ 0.58 & 26.43 $\pm$ 1.66 & 32.87 $\pm$ 0.12 \\
 montezuma revenge & 566.67 $\pm$ 235.70 & 1664.89 $\pm$ 1177.26 & \bf{11539.69 $\pm$ 1227.71} & 7619.70 $\pm$ 3444.76 & 7966.67 $\pm$ 2531.58 \\
 pitfall & 0.00 $\pm$ 0.00 & 0.00 $\pm$ 0.00 & 15195.27 $\pm$ 8005.22 & 2979.57 $\pm$ 2919.08 & \bf{16402.61 $\pm$ 10471.27} \\
 pong & \bf{21.00 $\pm$ 0.00} & 21.00 $\pm$ 0.00 & 21.00 $\pm$ 0.00 & 20.56 $\pm$ 0.28 & 21.00 $\pm$ 0.00 \\
 private eye & 21729.91 $\pm$ 9571.60 & 22480.31 $\pm$ 10362.99 & 63953.38 $\pm$ 26278.51 & 43823.40 $\pm$ 4808.23 & \bf{80581.86 $\pm$ 28331.16} \\
 skiing & -10784.13 $\pm$ 2539.27 & -4596.26 $\pm$ 601.04 & -19817.99 $\pm$ 7755.19 & \bf{-4051.99 $\pm$ 569.78} & -4278.86 $\pm$ 270.96 \\
 solaris & \bf{52500.89 $\pm$ 2910.14} & 14814.76 $\pm$ 11361.16 & 44771.13 $\pm$ 4920.53 & 43963.59 $\pm$ 5765.41 & 17254.14 $\pm$ 5840.70 \\
 surround & \bf{10.00 $\pm$ 0.00} & 10.00 $\pm$ 0.00 & 9.77 $\pm$ 0.23 & -7.57 $\pm$ 0.05 & 9.60 $\pm$ 0.20 \\
 venture & 2100.00 $\pm$ 0.00 & 1774.89 $\pm$ 83.79 & \bf{3249.01 $\pm$ 544.19} & 2228.04 $\pm$ 305.50 & 2576.98 $\pm$ 394.84 \\
\hline
\end{tabular}
\normalsize

\clearpage
\subsection{Backprop window length comparison}
\label{app:tracelength}
\begin{figure}[!ht]
    \centering
    \includegraphics[width=0.6\textwidth]{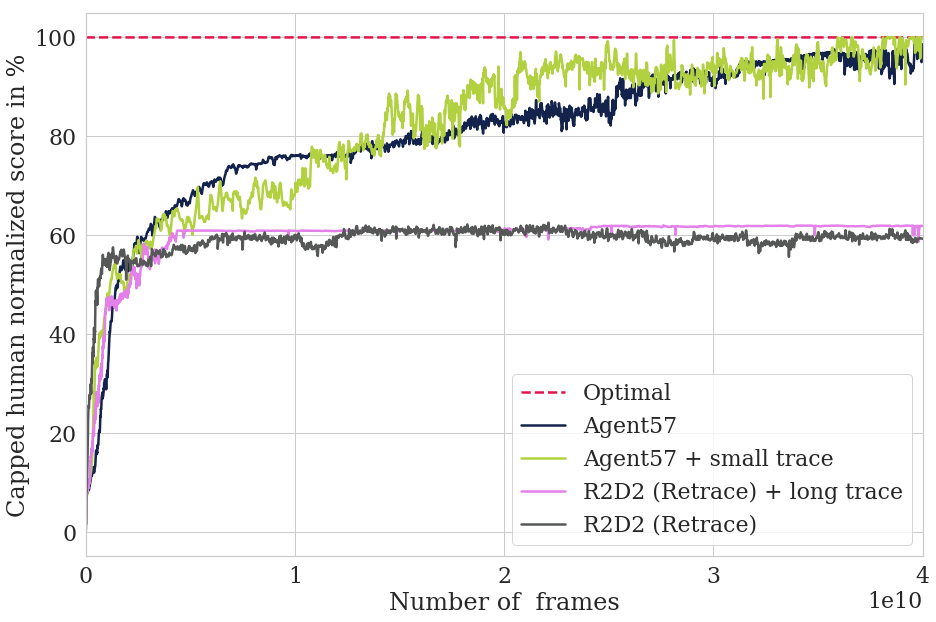}
    \caption{Performance comparison for short and long backprob window length on the 10-game \emph{challenging set}.}
\end{figure}

\subsection{Identity versus $h$-transform mixes comparison}
\label{app:mix}
\begin{figure}[!ht]
    \centering
    \includegraphics[width=0.6\textwidth]{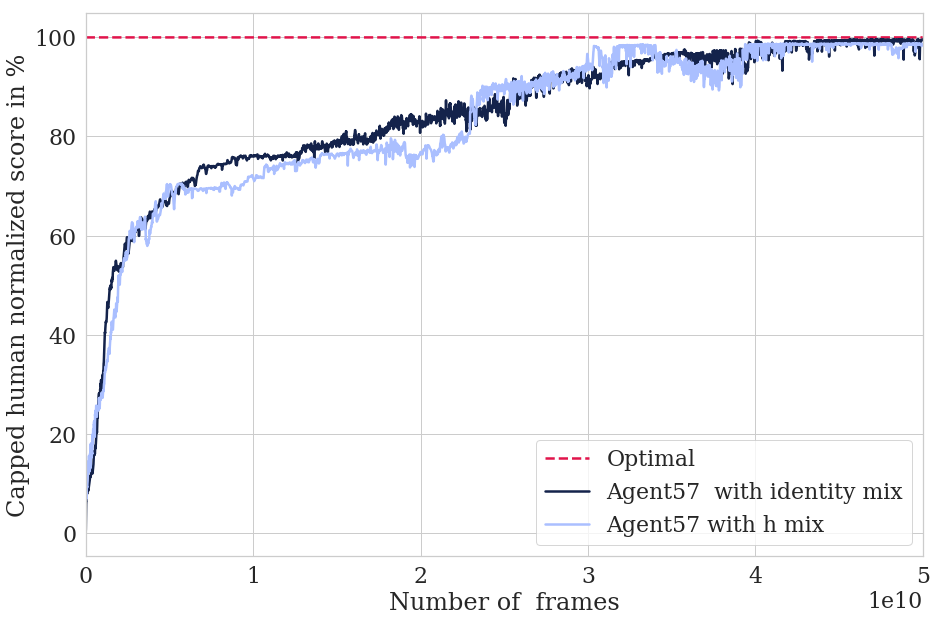}
    \label{fig:mix}
    \caption{Performance comparison for identity versus $h$-transform mixes on the 10-game \emph{challenging set}.}
\end{figure}
As shown in Fig~\ref{fig:mix}, choosing an identity or an $h$-transform mix does not seem to make a difference in terms of performance. The only real important thing is that a combination between extrinsic and intrinsic happens whether it is linear or not. In addition, one can remark that for extreme values of $\beta$ ($\beta = 0$, $\beta>>1$), the quantities $Q^e_k(x, a) + \beta Q^i_k(x, a)$ and  $h^{-1}(Q^e_k(x, a)) + \beta h^{-1}(Q^i_k(x, a))$ have the same $\argmax_{a\in\mathcal{A}}$ because $h^{-1}$ is strictly increasing. Therefore, this means that on the extremes values of $\beta$, the transform and normal value iteration schemes converge towards the same policy. For in between values of $\beta$, this is not the case. But we can conjecture that when a transform operator and and identity mix are used, the value iteration scheme approximates a state-action value function that is optimal with respect to a non-linear combination of the intrinsic and extrinsic rewards $r^i, r^e$, respectively. 

\subsection{Atari 57 Table of Scores}
\label{app:tab}
\small
\begin{tabular}{|c|c|c|c|c|c|}
\hline
 Games & Average Human & Random & Agent57 & R2D2 (Bandit) & MuZero \\
\hline
 alien & 7127.70 & 227.80 & 297638.17 $\pm$ 37054.55 & 464232.43 $\pm$ 7988.66 & \bf{741812.63} \\
 amidar & 1719.50 & 5.80 & 29660.08 $\pm$ 880.39 & \bf{31331.37 $\pm$ 817.79} & 28634.39 \\
 assault & 742.00 & 222.40 & 67212.67 $\pm$ 6150.59 & 110100.04 $\pm$ 346.06 & \bf{143972.03} \\
 asterix & 8503.30 & 210.00 & 991384.42 $\pm$ 9493.32 & \bf{999354.03 $\pm$ 12.94} & 998425.00 \\
 asteroids & 47388.70 & 719.10 & 150854.61 $\pm$ 16116.72 & 431072.45 $\pm$ 1799.13 & \bf{6785558.64} \\
 atlantis & 29028.10 & 12850.00 & 1528841.76 $\pm$ 28282.53 & 1660721.85 $\pm$ 14643.83 & \bf{1674767.20} \\
 bank heist & 753.10 & 14.20 & 23071.50 $\pm$ 15834.73 & \bf{27117.85 $\pm$ 963.12} & 1278.98 \\
 battle zone & 37187.50 & 2360.00 & 934134.88 $\pm$ 38916.03 & \bf{992600.31 $\pm$ 1096.19} & 848623.00 \\
 beam rider & 16926.50 & 363.90 & 300509.80 $\pm$ 13075.35 & 390603.06 $\pm$ 23304.09 & \bf{4549993.53} \\
 berzerk & 2630.40 & 123.70 & 61507.83 $\pm$ 26539.54 & 77725.62 $\pm$ 4556.93 & \bf{85932.60} \\
 bowling & 160.70 & 23.10 & 251.18 $\pm$ 13.22 & 161.77 $\pm$ 99.84 & \bf{260.13} \\
 boxing & 12.10 & 0.10 & 100.00 $\pm$ 0.00 & \bf{100.00 $\pm$ 0.00} & 100.00 \\
 breakout & 30.50 & 1.70 & 790.40 $\pm$ 60.05 & 863.92 $\pm$ 0.08 & \bf{864.00} \\
 centipede & 12017.00 & 2090.90 & 412847.86 $\pm$ 26087.14 & 908137.24 $\pm$ 7330.99 & \bf{1159049.27} \\
 chopper command & 7387.80 & 811.00 & 999900.00 $\pm$ 0.00 & \bf{999900.00 $\pm$ 0.00} & 991039.70 \\
 crazy climber & 35829.40 & 10780.50 & 565909.85 $\pm$ 89183.85 & \bf{729482.83 $\pm$ 87975.74} & 458315.40 \\
 defender & 18688.90 & 2874.50 & 677642.78 $\pm$ 16858.59 & 730714.53 $\pm$ 715.54 & \bf{839642.95} \\
 demon attack & 1971.00 & 152.10 & 143161.44 $\pm$ 220.32 & 143913.32 $\pm$ 92.93 & \bf{143964.26} \\
 double dunk & -16.40 & -18.60 & 23.93 $\pm$ 0.06 & \bf{24.00 $\pm$ 0.00} & 23.94 \\
 enduro & 860.50 & 0.00 & 2367.71 $\pm$ 8.69 & 2378.66 $\pm$ 3.66 & \bf{2382.44} \\
 fishing derby & -38.70 & -91.70 & 86.97 $\pm$ 3.25 & 90.34 $\pm$ 2.66 & \bf{91.16} \\
 freeway & 29.60 & 0.00 & 32.59 $\pm$ 0.71 & \bf{34.00 $\pm$ 0.00} & 33.03 \\
 frostbite & 4334.70 & 65.20 & 541280.88 $\pm$ 17485.76 & 309077.30 $\pm$ 274879.03 & \bf{631378.53} \\
 gopher & 2412.50 & 257.60 & 117777.08 $\pm$ 3108.06 & 129736.13 $\pm$ 653.03 & \bf{130345.58} \\
 gravitar & 3351.40 & 173.00 & 19213.96 $\pm$ 348.25 & \bf{21068.03 $\pm$ 497.25} & 6682.70 \\
 hero & 30826.40 & 1027.00 & \bf{114736.26 $\pm$ 49116.60} & 49339.62 $\pm$ 4617.76 & 49244.11 \\
 ice hockey & 0.90 & -11.20 & 63.64 $\pm$ 6.48 & \bf{86.59 $\pm$ 0.59} & 67.04 \\
 jamesbond & 302.80 & 29.00 & 135784.96 $\pm$ 9132.28 & \bf{158142.36 $\pm$ 904.45} & 41063.25 \\
 kangaroo & 3035.00 & 52.00 & \bf{24034.16 $\pm$ 12565.88} & 18284.99 $\pm$ 817.25 & 16763.60 \\
 krull & 2665.50 & 1598.00 & 251997.31 $\pm$ 20274.39 & 245315.44 $\pm$ 48249.07 & \bf{269358.27} \\
 kung fu master & 22736.30 & 258.50 & 206845.82 $\pm$ 11112.10 & \bf{267766.63 $\pm$ 2895.73} & 204824.00 \\
 montezuma revenge & 4753.30 & 0.00 & \bf{9352.01 $\pm$ 2939.78} & 3000.00 $\pm$ 0.00 & 0.00 \\
 ms pacman & 6951.60 & 307.30 & 63994.44 $\pm$ 6652.16 & 62595.90 $\pm$ 1755.82 & \bf{243401.10} \\
 name this game & 8049.00 & 2292.30 & 54386.77 $\pm$ 6148.50 & 138030.67 $\pm$ 5279.91 & \bf{157177.85} \\
 phoenix & 7242.60 & 761.40 & 908264.15 $\pm$ 28978.92 & \bf{990638.12 $\pm$ 6278.77} & 955137.84 \\
 pitfall & 6463.70 & -229.40 & \bf{18756.01 $\pm$ 9783.91} & 0.00 $\pm$ 0.00 & 0.00 \\
 pong & 14.60 & -20.70 & 20.67 $\pm$ 0.47 & \bf{21.00 $\pm$ 0.00} & 21.00 \\
 private eye & 69571.30 & 24.90 & \bf{79716.46 $\pm$ 29515.48} & 40700.00 $\pm$ 0.00 & 15299.98 \\
 qbert & 13455.00 & 163.90 & 580328.14 $\pm$ 151251.66 & \bf{777071.30 $\pm$ 190653.94} & 72276.00 \\
 riverraid & 17118.00 & 1338.50 & 63318.67 $\pm$ 5659.55 & 93569.66 $\pm$ 13308.08 & \bf{323417.18} \\
 road runner & 7845.00 & 11.50 & 243025.80 $\pm$ 79555.98 & 593186.78 $\pm$ 88650.69 & \bf{613411.80} \\
 robotank & 11.90 & 2.20 & 127.32 $\pm$ 12.50 & \bf{144.00 $\pm$ 0.00} & 131.13 \\
 seaquest & 42054.70 & 68.40 & 999997.63 $\pm$ 1.42 & \bf{999999.00 $\pm$ 0.00} & 999976.52 \\
 skiing & -4336.90 & -17098.10 & -4202.60 $\pm$ 607.85 & \bf{-3851.44 $\pm$ 517.52} & -29968.36 \\
 solaris & 12326.70 & 1236.30 & 44199.93 $\pm$ 8055.50 & \bf{67306.29 $\pm$ 10378.22} & 56.62 \\
 space invaders & 1668.70 & 148.00 & 48680.86 $\pm$ 5894.01 & 67898.71 $\pm$ 1744.74 & \bf{74335.30} \\
 star gunner & 10250.00 & 664.00 & 839573.53 $\pm$ 67132.17 & \bf{998600.28 $\pm$ 218.66} & 549271.70 \\
 surround & 6.50 & -10.00 & 9.50 $\pm$ 0.19 & \bf{10.00 $\pm$ 0.00} & 9.99 \\
 tennis & -8.30 & -23.80 & 23.84 $\pm$ 0.10 & \bf{24.00 $\pm$ 0.00} & 0.00 \\
 time pilot & 5229.20 & 3568.00 & 405425.31 $\pm$ 17044.45 & 460596.49 $\pm$ 3139.33 & \bf{476763.90} \\
 tutankham & 167.60 & 11.40 & \bf{2354.91 $\pm$ 3421.43} & 483.78 $\pm$ 37.90 & 491.48 \\
 up n down & 11693.20 & 533.40 & 623805.73 $\pm$ 23493.75 & 702700.36 $\pm$ 8937.59 & \bf{715545.61} \\
 venture & 1187.50 & 0.00 & \bf{2623.71 $\pm$ 442.13} & 2258.93 $\pm$ 29.90 & 0.40 \\
 video pinball & 17667.90 & 0.00 & 992340.74 $\pm$ 12867.87 & \bf{999645.92 $\pm$ 57.93} & 981791.88 \\
 wizard of wor & 4756.50 & 563.50 & 157306.41 $\pm$ 16000.00 & 183090.81 $\pm$ 6070.10 & \bf{197126.00} \\
 yars revenge & 54576.90 & 3092.90 & 998532.37 $\pm$ 375.82 & \bf{999807.02 $\pm$ 54.85} & 553311.46 \\
 zaxxon & 9173.30 & 32.50 & 249808.90 $\pm$ 58261.59 & 370649.03 $\pm$ 19761.32 & \bf{725853.90} \\
\hline
\end{tabular}

\begin{tabular}{|c|c|c|c|c|}
\hline
 Games & Agent57 & NGU & R2D2 (Retrace) & R2D2 \\
\hline
 alien & 297638.17 $\pm$ 37054.55 & 312024.15 $\pm$ 91963.92 & 228483.74 $\pm$ 111660.11 & \bf{399709.08 $\pm$ 106191.42} \\
 amidar & 29660.08 $\pm$ 880.39 & 18369.47 $\pm$ 2141.76 & 28777.05 $\pm$ 803.90 & \bf{30338.91 $\pm$ 1087.62} \\
 assault & 67212.67 $\pm$ 6150.59 & 42829.17 $\pm$ 7452.17 & 46003.71 $\pm$ 8996.65 & \bf{124931.33 $\pm$ 2627.16} \\
 asterix & 991384.42 $\pm$ 9493.32 & 996141.15 $\pm$ 3993.26 & 998867.54 $\pm$ 191.35 & \bf{999403.53 $\pm$ 76.75} \\
 asteroids & 150854.61 $\pm$ 16116.72 & 248951.23 $\pm$ 7561.86 & 345910.03 $\pm$ 13189.10 & \bf{394765.73 $\pm$ 16944.82} \\
 atlantis & 1528841.76 $\pm$ 28282.53 & \bf{1659575.47 $\pm$ 4140.68} & 1659411.83 $\pm$ 9934.57 & 1644680.76 $\pm$ 5784.97 \\
 bank heist & 23071.50 $\pm$ 15834.73 & 20012.54 $\pm$ 20377.89 & 16726.07 $\pm$ 10992.11 & \bf{38536.66 $\pm$ 11645.73} \\
 battle zone & 934134.88 $\pm$ 38916.03 & 813965.40 $\pm$ 94503.50 & 845666.67 $\pm$ 51527.68 & \bf{956179.17 $\pm$ 31019.66} \\
 beam rider & \bf{300509.80 $\pm$ 13075.35} & 75889.70 $\pm$ 18226.52 & 123281.81 $\pm$ 4566.16 & 246078.69 $\pm$ 3667.61 \\
 berzerk & 61507.83 $\pm$ 26539.54 & 45601.93 $\pm$ 5170.98 & \bf{73475.91 $\pm$ 8107.24} & 64852.56 $\pm$ 17875.17 \\
 bowling & 251.18 $\pm$ 13.22 & 215.38 $\pm$ 13.27 & \bf{257.88 $\pm$ 4.84} & 229.39 $\pm$ 24.57 \\
 boxing & \bf{100.00 $\pm$ 0.00} & 99.71 $\pm$ 0.25 & 100.00 $\pm$ 0.00 & 99.27 $\pm$ 0.35 \\
 breakout & 790.40 $\pm$ 60.05 & 625.86 $\pm$ 42.66 & 859.60 $\pm$ 2.04 & \bf{863.25 $\pm$ 0.34} \\
 centipede & 412847.86 $\pm$ 26087.14 & 596427.16 $\pm$ 7149.84 & \bf{737655.85 $\pm$ 25568.85} & 693733.73 $\pm$ 74495.81 \\
 chopper command & 999900.00 $\pm$ 0.00 & 999900.00 $\pm$ 0.00 & 999900.00 $\pm$ 0.00 & \bf{999900.00 $\pm$ 0.00} \\
 crazy climber & \bf{565909.85 $\pm$ 89183.85} & 351390.64 $\pm$ 62150.96 & 322741.20 $\pm$ 23024.88 & 549054.89 $\pm$ 39413.08 \\
 defender & 677642.78 $\pm$ 16858.59 & 684414.06 $\pm$ 3876.41 & 681291.73 $\pm$ 3469.95 & \bf{692114.71 $\pm$ 4864.99} \\
 demon attack & 143161.44 $\pm$ 220.32 & 143695.73 $\pm$ 154.88 & \bf{143899.22 $\pm$ 53.78} & 143830.91 $\pm$ 107.18 \\
 double dunk & 23.93 $\pm$ 0.06 & -12.63 $\pm$ 5.29 & \bf{24.00 $\pm$ 0.00} & 23.97 $\pm$ 0.03 \\
 enduro & 2367.71 $\pm$ 8.69 & 2095.40 $\pm$ 80.81 & 2372.77 $\pm$ 3.50 & \bf{2380.22 $\pm$ 5.47} \\
 fishing derby & 86.97 $\pm$ 3.25 & 34.62 $\pm$ 4.91 & \bf{87.83 $\pm$ 2.78} & 87.81 $\pm$ 1.28 \\
 freeway & 32.59 $\pm$ 0.71 & 28.71 $\pm$ 2.07 & \bf{33.48 $\pm$ 0.16} & 32.90 $\pm$ 0.11 \\
 frostbite & \bf{541280.88 $\pm$ 17485.76} & 284044.19 $\pm$ 227850.49 & 12290.11 $\pm$ 7936.49 & 446703.01 $\pm$ 63780.51 \\
 gopher & 117777.08 $\pm$ 3108.06 & 119110.87 $\pm$ 463.03 & 119803.94 $\pm$ 3197.88 & \bf{126241.97 $\pm$ 519.70} \\
 gravitar & \bf{19213.96 $\pm$ 348.25} & 14771.91 $\pm$ 843.17 & 14194.45 $\pm$ 1250.63 & 17352.78 $\pm$ 2675.27 \\
 hero & \bf{114736.26 $\pm$ 49116.60} & 71592.84 $\pm$ 12109.10 & 54967.97 $\pm$ 5411.73 & 39786.01 $\pm$ 7638.19 \\
 ice hockey & 63.64 $\pm$ 6.48 & -3.15 $\pm$ 0.47 & 86.56 $\pm$ 1.21 & \bf{86.89 $\pm$ 0.88} \\
 jamesbond & \bf{135784.96 $\pm$ 9132.28} & 28725.27 $\pm$ 2902.52 & 32926.31 $\pm$ 3073.94 & 28988.32 $\pm$ 263.79 \\
 kangaroo & 24034.16 $\pm$ 12565.88 & \bf{37392.82 $\pm$ 6170.95} & 15185.87 $\pm$ 931.58 & 14492.75 $\pm$ 5.29 \\
 krull & 251997.31 $\pm$ 20274.39 & 150896.04 $\pm$ 33729.56 & 149221.98 $\pm$ 17583.30 & \bf{291043.06 $\pm$ 10051.59} \\
 kung fu master & 206845.82 $\pm$ 11112.10 & 215938.95 $\pm$ 22050.67 & 228228.90 $\pm$ 5316.74 & \bf{252876.65 $\pm$ 10424.57} \\
 montezuma revenge & 9352.01 $\pm$ 2939.78 & \bf{19093.74 $\pm$ 12627.66} & 2300.00 $\pm$ 668.33 & 2666.67 $\pm$ 235.70 \\
 ms pacman & \bf{63994.44 $\pm$ 6652.16} & 48695.12 $\pm$ 1599.94 & 45011.73 $\pm$ 1822.30 & 50337.02 $\pm$ 4004.55 \\
 name this game & 54386.77 $\pm$ 6148.50 & 25608.90 $\pm$ 1943.41 & 74104.70 $\pm$ 9053.70 & \bf{74501.48 $\pm$ 11562.26} \\
 phoenix & 908264.15 $\pm$ 28978.92 & \bf{966685.41 $\pm$ 6127.24} & 937874.90 $\pm$ 22525.79 & 876045.70 $\pm$ 25511.04 \\
 pitfall & \bf{18756.01 $\pm$ 9783.91} & 15334.30 $\pm$ 15106.90 & -0.45 $\pm$ 0.50 & 0.00 $\pm$ 0.00 \\
 pong & 20.67 $\pm$ 0.47 & 19.85 $\pm$ 0.31 & 20.95 $\pm$ 0.01 & \bf{21.00 $\pm$ 0.00} \\
 private eye & 79716.46 $\pm$ 29515.48 & \bf{100314.44 $\pm$ 291.22} & 34601.01 $\pm$ 5266.39 & 18765.05 $\pm$ 16672.27 \\
 qbert & 580328.14 $\pm$ 151251.66 & 479024.20 $\pm$ 98094.39 & 434753.72 $\pm$ 99793.58 & \bf{771069.21 $\pm$ 152722.56} \\
 riverraid & \bf{63318.67 $\pm$ 5659.55} & 40770.82 $\pm$ 748.42 & 43174.10 $\pm$ 2335.12 & 54280.32 $\pm$ 1245.60 \\
 road runner & 243025.80 $\pm$ 79555.98 & 151326.54 $\pm$ 77209.43 & 116149.17 $\pm$ 18257.21 & \bf{613659.42 $\pm$ 397.72} \\
 robotank & 127.32 $\pm$ 12.50 & 11.62 $\pm$ 0.67 & \bf{143.59 $\pm$ 0.29} & 130.72 $\pm$ 9.75 \\
 seaquest & 999997.63 $\pm$ 1.42 & \bf{999999.00 $\pm$ 0.00} & 999999.00 $\pm$ 0.00 & 999999.00 $\pm$ 0.00 \\
 skiing & \bf{-4202.60 $\pm$ 607.85} & -24271.33 $\pm$ 6936.26 & -14576.05 $\pm$ 875.96 & -17797.59 $\pm$ 866.55 \\
 solaris & \bf{44199.93 $\pm$ 8055.50} & 7254.03 $\pm$ 3653.55 & 6566.03 $\pm$ 2209.91 & 11247.88 $\pm$ 1999.22 \\
 space invaders & 48680.86 $\pm$ 5894.01 & 48087.13 $\pm$ 11219.39 & 36069.75 $\pm$ 23408.12 & \bf{67229.37 $\pm$ 2316.31} \\
 star gunner & 839573.53 $\pm$ 67132.17 & 450096.08 $\pm$ 158979.59 & 420337.48 $\pm$ 8309.08 & \bf{923739.89 $\pm$ 69234.32} \\
 surround & 9.50 $\pm$ 0.19 & -9.32 $\pm$ 0.67 & 9.96 $\pm$ 0.01 & \bf{10.00 $\pm$ 0.00} \\
 tennis & 23.84 $\pm$ 0.10 & 11.06 $\pm$ 6.10 & \bf{24.00 $\pm$ 0.00} & 7.93 $\pm$ 11.36 \\
 time pilot & 405425.31 $\pm$ 17044.45 & 368520.34 $\pm$ 70829.26 & 452966.67 $\pm$ 5300.62 & \bf{454055.63 $\pm$ 2205.07} \\
 tutankham & \bf{2354.91 $\pm$ 3421.43} & 197.90 $\pm$ 7.47 & 466.59 $\pm$ 38.40 & 413.80 $\pm$ 3.89 \\
 up n down & 623805.73 $\pm$ 23493.75 & 630463.10 $\pm$ 31175.20 & \bf{679303.61 $\pm$ 4852.85} & 599134.12 $\pm$ 3394.48 \\
 venture & \bf{2623.71 $\pm$ 442.13} & 1747.32 $\pm$ 101.40 & 2013.31 $\pm$ 11.24 & 2047.51 $\pm$ 20.83 \\
 video pinball & 992340.74 $\pm$ 12867.87 & 973898.32 $\pm$ 20593.14 & 964670.12 $\pm$ 4015.52 & \bf{999697.05 $\pm$ 53.37} \\
 wizard of wor & 157306.41 $\pm$ 16000.00 & 121791.35 $\pm$ 27909.14 & 134017.82 $\pm$ 11871.88 & \bf{179376.15 $\pm$ 6659.14} \\
 yars revenge & 998532.37 $\pm$ 375.82 & 997642.09 $\pm$ 455.73 & 998474.20 $\pm$ 589.50 & \bf{999748.54 $\pm$ 46.19} \\
 zaxxon & 249808.90 $\pm$ 58261.59 & 129330.99 $\pm$ 56872.31 & 114990.68 $\pm$ 56726.18 & \bf{366028.59 $\pm$ 49366.03} \\
\hline
\end{tabular}
\normalsize
\clearpage
\subsection{Atari 57 Learning Curves}
\label{app:scores}
\begin{figure}[!ht]
    \centering
    \includegraphics[width=1.0\textwidth]{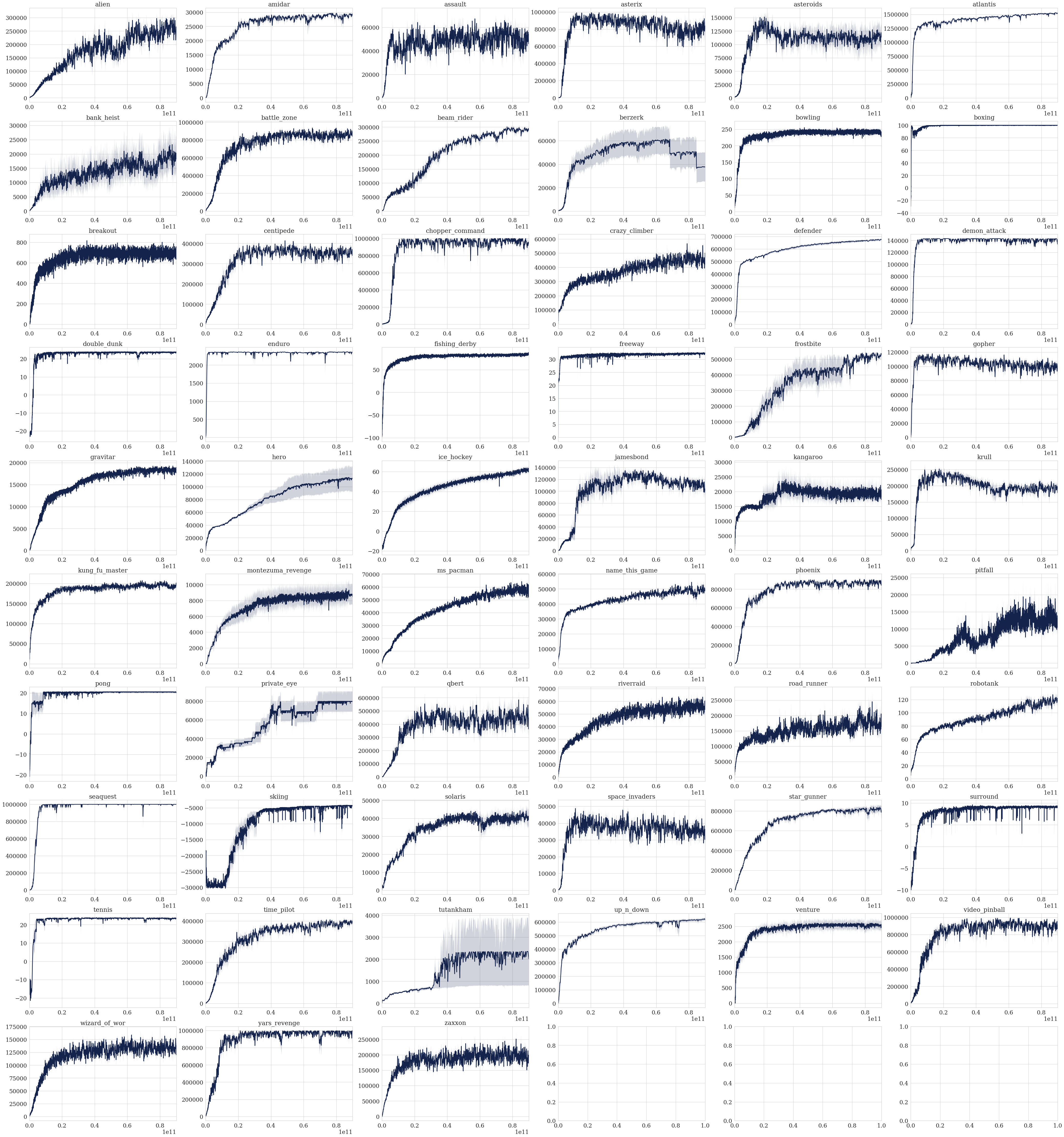}
    \caption{Learning curves for Agent57 on Atari57.} 
    \label{fig:atari57learningcurves}
\end{figure}

\newpage
\subsection{Videos}
\label{app:videos}

We provide several videos in \url{https://sites.google.com/corp/view/agent57}. We show 

\begin{itemize}

\item {\bf Agent57 on all 57 games:} We provide an example video for each game in the Atari 57 sweep in which Agent57 surpasses the human baseline.

\item {\bf State-action Value Function Parameterization:} To illustrate the importance of the value function parametrization we show videos in two games \textit{Ice Hockey} and \textit{Surround}. We show videos for exploitative and exploratory policies for both NGU and Agent57. In \textit{Ice Hockey}, exploratory and exploitative policies are quite achieving very different scores. Specifically the exploratory policy does not aim to score goals, it prefers to move around the court exploring new configurations. On the other hand, NGU with a single architecture is unable to learn both policies simultaneously, while Agent57 show very diverse performance. In the case of \textit{Surround} NGU is again unable to learn. We conjecture that the exploratory policy chooses to loose a point in order to start afresh increasing the diversity of the observations. Agent57 is able to overcome this problem and both exploitative and exploratory policies are able to obtain scores surpassing the human baseline.

\item {\bf Adaptive Discount Factor:} We show example videos for R2D2 (bandit) and R2D2 (retrace) in the game \textit{James Bond}. R2D2 (retrace) learns to clear the game with a final score in the order of 30,000 points. R2D2 (bandit) in contrast, learns to delay the end of the game to collect significantly more rewards with a score around 140,000 points. To achieve this, the adaptive mechanism in the meta-controller, selects policies with very high discount factors.

\item {\bf Backprop Through Time Window Size:} We provide videos showing example episodes for NGU and Agent57 on the game of \textit{Solaris}. In order to achieve high scores, the agent needs to learn to move around the grid screen and look for enemies. This is a long term credit assignment problem as the agent needs to bind the actions taken on the grid screen with the reward achieved many time steps later.

\end{itemize}

\end{document}